\documentclass[runningheads]{llncs}
\usepackage[utf8]{inputenc}
\usepackage{scalerel,graphicx,xparse}
\usepackage[tbtags]{amsmath}
\usepackage{cite}
\usepackage{amssymb}
\usepackage{booktabs}
\usepackage{color}
\usepackage{xcolor}
\usepackage{dsfont}
\usepackage{mathtools}
\usepackage{multirow}
\usepackage{wrapfig}
\usepackage{enumitem}
\usepackage{listings}
\usepackage{caption}
\usepackage[labelformat=simple]{subcaption}
\usepackage[noend]{algpseudocode}
\usepackage{multicol} 
\usepackage{hyperref}
\usepackage{placeins}
\usepackage[misc]{ifsym}
\usepackage[bottom]{footmisc}
\definecolor{codegreen}{rgb}{0,0.6,0}
\definecolor{codegray}{rgb}{0.5,0.5,0.5}
\definecolor{codepurple}{rgb}{0.58,0,0.82}
\definecolor{backcolour}{rgb}{0.95,0.95,0.92}

\lstdefinestyle{pythonstyle}{
  backgroundcolor=\color{backcolour}, commentstyle=\color{codegreen},
  keywordstyle=\color{magenta},
  numberstyle=\scriptsize\color{codegray},
  stringstyle=\color{codepurple},
  basicstyle=\ttfamily\footnotesize,
  breakatwhitespace=false,         
  breaklines=true,                 
  captionpos=b,                    
  keepspaces=true,                 
  numbers=left,                    
  numbersep=5pt,                  
  showspaces=false,                
  showstringspaces=false,
  showtabs=false,                  
  tabsize=2
}

\newcommand{\eg}{\emph{e.g.}~} 

\newcommand{\ie}{\emph{i.e.}~} 

\newcommand{\cf}{\emph{cf.}~}

\definecolor{blue}{rgb}{0.3, 0.3, 0.9}

\definecolor{green}{rgb}{0.1, 0.5, 0.1}

\definecolor{orange}{rgb}{1, 0.5, 0}
 
\definecolor{modif}{rgb}{0.3, 0.7, 0.1}

\newcommand{\round}[1]{\ensuremath{\lfloor#1\rceil}}

\usepackage{geometry}
\newcommand\blfootnote[1]{%
  \begingroup
  \renewcommand\thefootnote{}\footnote{\color{blue}#1}%
  \addtocounter{footnote}{-1}%
  \endgroup
}

\begin{document}
\toctitle{Boosting Object Representation Learning \\ via Motion and Object Continuity}
\title{Boosting Object Representation Learning \\ via Motion and Object Continuity}
%
%
\author{Quentin Delfosse\inst{1,*} [\Letter] \and 
Wolfgang Stammer\inst{1,2,*} \and 
Thomas Rothenbächer\inst{1} \and \\
Dwarak Vittal\inst{1} \and 
Kristian Kersting\inst{1,2,3,4}}
\authorrunning{Q. Delfosse et al.}
%
\institute{AI \& ML Lab, TU Darmstadt, Germany \and
Hessian Center for AI (hessian.AI), Darmstadt, Germany \and 
German Research Center for AI (DFKI) \and
Centre for Cognitive Science, TU Darmstadt, Germany\\
Correspondence to: \href{mailto:quentin.delfosse@cs.tu-darmstadt.de}{quentin.delfosse@cs.tu-darmstadt.de}}
%

\toctitle{Boosting Object Representation Learning via Motion and Object Continuity}
\tocauthor{Quentin~Delfosse, Wolfgang~Stammer, Thomas~Rothenbächer, Dwarak~Vittal, Kristian~Kersting}

\maketitle              
\begin{abstract}
Recent unsupervised multi-object detection models have shown impressive performance improvements, largely attributed to novel architectural inductive biases. Unfortunately, despite their good object localization and segmentation capabilities, their object encodings may still be suboptimal for downstream reasoning tasks, such as reinforcement learning.  
To overcome this, we propose to exploit object motion and continuity (objects do not pop in and out of existence). This is accomplished through two mechanisms: (i) providing temporal loss-based priors on object locations, and (ii) a contrastive object continuity loss across consecutive frames. Rather than developing an explicit deep architecture, the resulting unsupervised Motion and Object Continuity (MOC) training scheme can be instantiated using any object detection model baseline.
Our results show large improvements in the performances of variational and slot-based models in terms of object discovery, convergence speed and overall latent object representations, particularly for playing Atari games. 
Overall, we show clear benefits of integrating motion and object continuity for downstream reasoning tasks, moving beyond object representation learning based only on reconstruction as well as evaluation based only on instance segmentation quality.\blfootnote{
Accepted at ECML PKDD 2023.\\ \textcolor{black}{ \qquad  $^*$These authors share equal contribution.}}
\end{abstract}

\begingroup 
\let\clearpage\relax
\section{Introduction}

Our surroundings largely consist of objects and their relations. In fact, decomposing the world in terms of objects is considered an important property of human perception and reasoning.
This insight has recently influenced a surge of research articles in the field of deep learning (DL), resulting in novel neural architectures and inductive biases for unsupervisedly decomposing a visual scene into objects (\eg \cite{eslami2016air, stelzner2019spair, SPACE2020, burgess2019monet, greff2019iodine, engelcke2020genesis, locatello2020slotattention, kipf2021SaVi}). Integrating these modules into systems for downstream reasoning tasks, \eg playing Atari games in reinforcement learning (RL) settings 
is a promising next step for human-centric AI (\cf Fig.~\ref{fig:oc_reasoner}). This integration could provide benefits both in overall performance and in terms of trustworthy human-machine interactions \cite{kambhampati2022symbols, stammer2021right}.

\begin{wrapfigure}{r}{0.455\linewidth}
    \centering
    \vspace{-0.65cm}
    \includegraphics[width=1.\linewidth]{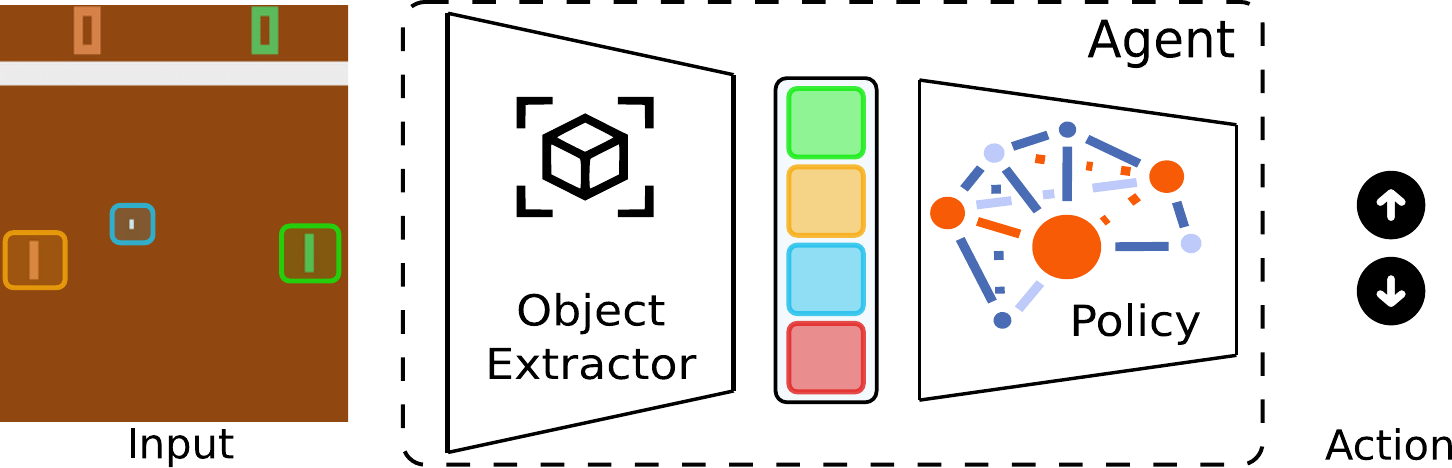}
    \caption{\textbf{An object-centric reasoner playing Pong.} The agent first extracts the object representation and then reasons on them to select an optimal action.}
    \vspace{-0.7cm}
  \label{fig:oc_reasoner} 
\end{wrapfigure}

However, although many of the previously mentioned works motivate their object discovery methods with the advantage of object-centric learning for complex downstream tasks, none of these explicitly optimize the object encodings for anything other than reconstruction. In fact, investigations of the object encodings of SPACE~\cite{SPACE2020}, SOTA on object detection for Atari games, reveal two shortcomings.  
First, although object detection does work for some games (\eg Space Invaders), it shows significant issues on other, even as simple as Pong (\cf Fig.~\ref{fig:motivation} for qualitative examples (top) and quantitative F-score evaluations (left)). 
Second, even for such a simple downstream task as object classification, its object encodings appear suboptimal even for games with high detection scores. This is exhibited by the accuracies of a ridge regression model~\cite{hoerl1970ridge1, hoerl1970ridge2} trained on SPACE's encodings for object classification (Fig.~\ref{fig:motivation} (left)). 
The encodings, mapped into a two-dimensional t-SNE~\cite{vandermaaten2008visualizing} embedding and colored by their ground truth class labels (\cf Fig.~\ref{fig:motivation} (right)), suggest a cluttered latent space. 

These results indicate open issues from two types of errors: (\textbf{Type I}) failures in object detection per se and (\textbf{Type II}) sub-optimal object representations.
Arguably, Type I is somewhat independent of Type II, as an optimal encoding is not a necessity for detecting objects in the first place. However, the Type II error is dependent on Type I, as an object representation can only exist for detected objects. Before integrating such recent modules into more general systems for reasoning tasks, we first need to tackle the remaining issues. 

We therefore propose a novel model-agnostic, self-supervised training scheme for improving object representation learning, particularly for integration into downstream reasoning tasks. We refer to this scheme as Motion and Object Continuity supervision (MOC). 
MOC jointly tackles Type I and II errors by incorporating the inductive biases of object motion (\ie objects tend to move in the real-world) and object continuity (\ie objects tend to still exist over time and do not pop in and out of existence) into object discovery models. 

\begin{figure*}[t!]
    \centerline{\includegraphics[width=.9\textwidth]{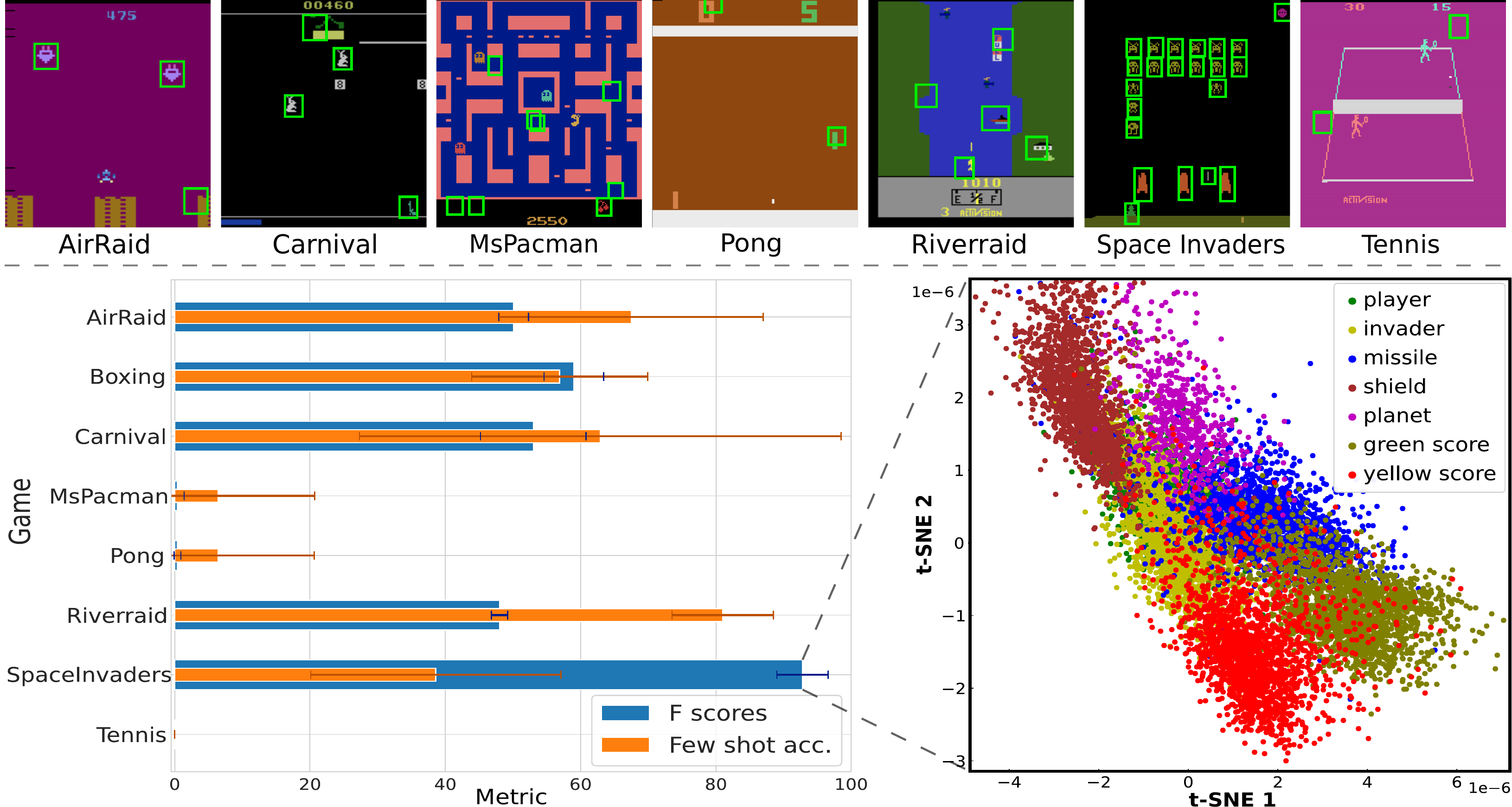}}
    \caption{Motivational example: unsupervised object detection models are insufficient for downstream tasks such as classification, exemplified here via SPACE~\cite{SPACE2020} on Atari environments. Top: Example images of SPACE detecting objects on different Atari games. Left: F-score for object detection (blue) and few shot classification accuracy of object encodings via ridge regression (orange, 64 objects per class, 0\%  accuracy corresponds to no object detected). Right: Two-dimensional t-SNE embedding of object encodings produced by SPACE for Space Invaders. 
    }
  \label{fig:motivation} 
\end{figure*}

It is based on the notion that visual input for humans is mainly perceived as image sequences. These contain rich object-based information signals, such as that objects tend to move over time, but also that an object identified in one frame will likely be present in the consecutive ones. We refer to the first property as 
\textit{object motion} (M) and the second property as \textit{object continuity} (OC). The concept of object continuity can be considered an extension of the Gestalt law of continuity~\cite{wertheimer1923untersuchungen} to the temporal dimension and has been identified to be an important property of infant and adult visual perception \cite{strickland2015principles}. 
MOC specifically makes use of this underlying information via two mechanisms. The first mechanism produces accurate object location priors based on estimated optical flow. The second presents a contrastive loss term on the object encodings of consecutive images.
Importantly, MOC is model-agnostic (\ie it can incorporate any object discovery model). 

In our experimental evaluations, we show the benefits of the novel MOC training scheme for object-centric learning for downstream reasoning tasks. For this, we integrate the models SPACE~\cite{SPACE2020} and Slot Attention~\cite{locatello2020slotattention} into MOC and quantify the benefits of MOC over the base models through a variety of metrics that highlight important properties of the training scheme. Importantly, and in contrast to previous works, we show the improved quality of MOC trained object encodings for downstream reasoning task performances such as playing Atari games and few-shot object classification.
In summary, we show that \textbf{inductive biases extracted from motion and object continuity are enough to boost object encodings of a predisposed object discovery model for downstream reasoning tasks}. 

Overall, our contributions are the following:
(i) We identify two error sources of object discovery models, hindering their current integration into modules for downstream reasoning tasks.
(ii) Introduce motion \textit{and} object continuity to DL as a novel self-supervised training scheme.
(iii) 
Create a novel open-source dataset, Atari-OCTA,  to train and evaluate object detection models on Atari games.
(iv) Empirically show 
on the SPACE and Slot Attention architectures that motion and object continuity greatly boosts downstream task performance for 
object-centric DL and allow for object-centric RL. 






We proceed as follows. 
We first give a brief introduction to object discovery methods, then explain our MOC scheme, detailing both Motion and Object Continuity supervision. We experimentally evaluate improvements on object detection and representations. Before concluding, we touch upon related work.

\section{Motion and Object Continuity}
\label{sec:methods}

The Motion and Object Continuity (MOC) training scheme utilizes two object properties for improving object detection and representation, both of which are required for good object representation learning for reasoning tasks and both of which can be discovered in visual observations over time. 
The first property corresponds to the fact that objects tend to move in an environment. Via optical flow estimations~\cite{teed2020raft, stone2021smurf} this can be translated to a motion detection mechanism that guides object localization and transforms unsupervised object detection into a ``motion supervised'' one. 
The second property describes the observation that objects exist at consecutive time points in proximity to their initial location and do not simply disappear. This can be integrated into an object continuity constraint, enforcing encoding alignment of objects over time. 

Our approach allows integrating any out-of-the-box object discovery method. For this work we have focused on providing MOC supervision to slot-based models \cite{locatello2020slotattention} and to models that separately encode the background and the foreground via separate variational autoencoders (VAE) (\eg \cite{SPACE2020, greff2019iodine}), where we specifically focus on SPACE~\cite{SPACE2020} due to its SOTA object discovery performance on images from Atari games.
We provide an overview of these methods in Fig.~7 
(\cf App.~B 
) and introduce the mathematical modelling of SPACE in the App.~B.1 
and of Slot Attention in the App.~B.2 
. 
In the following, we present a high-level description of MOC supervision, illustrated in Fig.~\ref{fig:time_supervision}, independent of the base models implementation. MOC implementation details for each model can be found in App.~C.2 
and C.3
.
Let us first provide some details on notations.

\begin{figure}[t]
    \center
    \includegraphics[width=.91\textwidth]{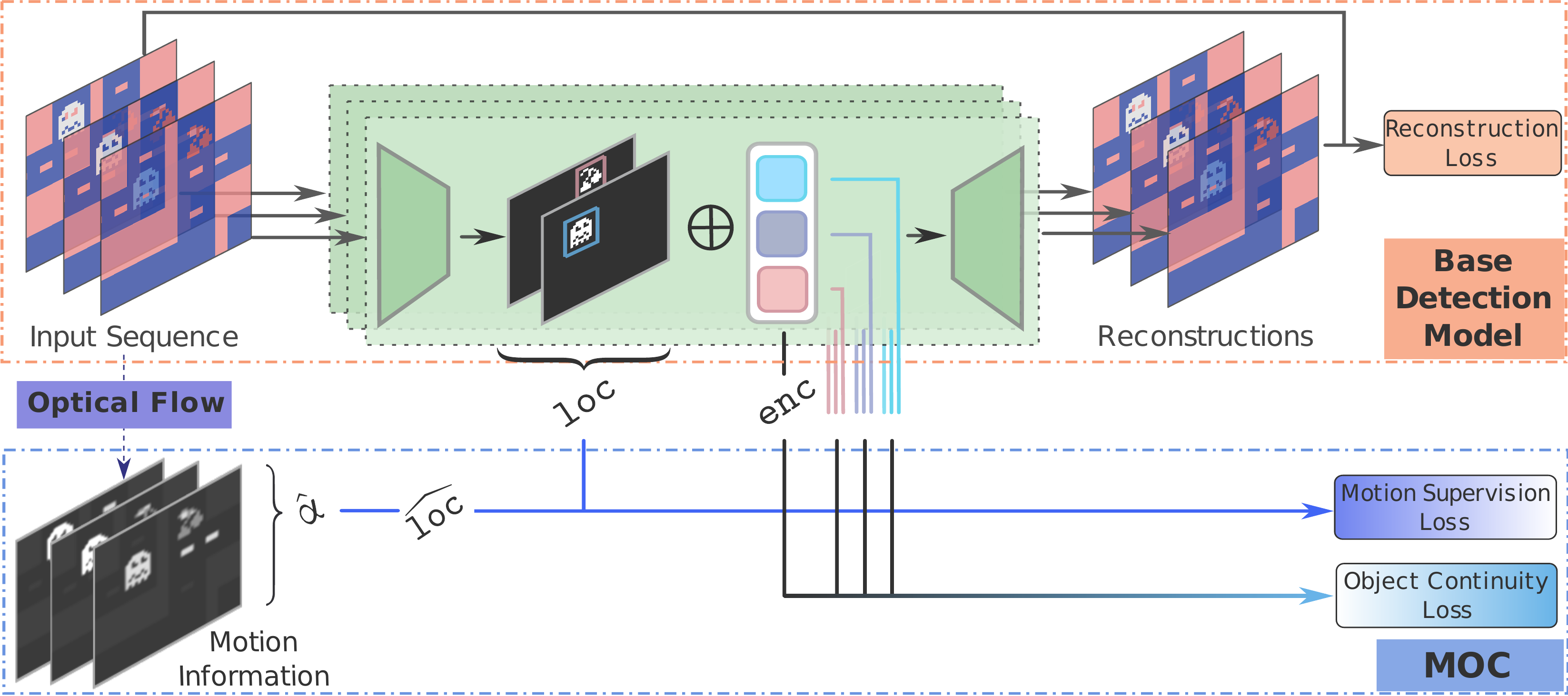}
    \caption{An overview of the MOC training scheme applied to a base object detection model, which provides location and object representations. In our MOC training scheme, (i) motion information (dark blue), is extracted from each frame, allowing to detect objects and directly update the model's latent location variables (\texttt{loc}). (ii) Object continuity (black + cyan) aligns the encodings (\texttt{enc}) of spatially close objects of consecutive frames using a contrastive loss.}
  \label{fig:time_supervision} 
\end{figure}

In the following we denote $\texttt{enc}$ as the object encoding obtained from a base model, \eg the slot encodings for Slot Attention~\cite{locatello2020slotattention}, or the $z_{what}$ encodings for SPACE~\cite{SPACE2020}. We further denote \texttt{loc} as the positional representation of an object.
Specifically, SPACE divides an image into a grid, and explicitly models the presence of an object in each cell using a $z_{pres}$ variable, and its position using a $z_{where}$ variable. For each object, $\texttt{loc}$ can thus be obtained using the $z_{where}$ variable of each cell in which an object is present (\eg $z_{pres} > 0.5$). 
For Slot Attention based models, on the other hand, the position information is implicitely contained within $\texttt{enc}$. 
However, the corresponding attention masks $\alpha$ also correspond to valid representations of object locations. We therefore denote these masks as the $\texttt{loc}$ variables of Slot Attention in the MOC framework. 
Furthermore, we denote $\mathcal{L}^\text{B}$ to represent the original training loss function of the base model, \eg reconstruction loss.

\subsection{Motion supervision}\label{sec:motion}
Let us now consider sequences of $T$ frames $\{\textbf{x}_t\}_{t=1}^T$, whereby $\textbf{x}_t$ corresponds to an RGB image ($\textbf{x}_t \in \mathbb{R}^{3 \times h \times w}$), at time step $t$. Given such a sequence of images, MOC requires preprocessing via any suitable optical flow estimation method (\eg \cite{anthwal2019overview, ZhengNLXLWL22DIP, BaiGSK22DEOFE,JeongLPK22OFC}) which accurately detects the motion of objects for the given domain and provides us with sufficient binary masks $\hat{\alpha}$ of the moving objects from which location information, $\widehat{\texttt{loc}}$, can be obtained for each object.

MOC now integrates these masks to provide feedback on the locations of initial object representations.
Specifically, for each frame $\textbf{x}_t$ we compute: 
\begin{equation}\label{eq:motion}
\mathcal{L}^\text{M}(\textbf{x}_t) := \sum_{i=1}^{N} \mathcal{L}^\text{M*}(\textbf{x}_t, \hat{\alpha}_t),
\end{equation}
where $\mathcal{L}^\text{M*}$ refers to the exact implementation for the base model including a weighting hyperparameter (\cf App.~C.2 
and C.3 
for SPACE and Slot Attention, respectively). 
In short, the masks obtained from optical flow ($\hat{\alpha}_t$) allow for direct supervision on the internal masks representation of Slot Attention based models. They allow us to construct $\widehat{\texttt{loc}}$ variables, to supervise $z_{pres}$ and $z_{where}$ used by SPACE models.

\subsection{Object Continuity}

Given that an object in consecutive image sequences tends to change its location only gradually and does not pop in and out of existence, we can integrate such bias by matching objects of similar position and latent encoding over consecutive frames. Particularly, we can make use of this information for explicitly improving the encoding space through a contrastive loss. 

For detected entities in consecutive frames, we apply the object continuity loss, $\mathcal{L}^\text{OC}$, on the internal representations (\ie $\texttt{enc}$) of the model. $\mathcal{L}^\text{OC}$ makes the representations of the same object depicted over several frames similar, and of different objects heterogeneous. We estimate whether two objects in consecutive frames represent the same entity based on their \texttt{loc} and \texttt{enc} variables.

In detail, we denote $\Omega_{t} = \{{o}_{t}^j\}_{j=1}^{c_t}$ the set of object representations (${o}_{t}^j = (\texttt{enc}_{t}^j, \texttt{loc}_{t}^j)$) that correspond to the $c_t$ detected objects in $\textbf{x}_t$.
Let $\texttt{enc}^*({o}_{t}^j)$ denote a function providing the object in $\Omega_{t\!+\!1}$ with the most similar encoding to ${o}_{t}^j$ based on the cosine similarity ($S_C$). Correspondingly, $\texttt{loc}^*({o}_{t}^j)$ is a function that provides the nearest object of $\Omega_{t\!+\!1}$ to ${o}_{t}^j$ based on the locations (\eg via Euclidean distance).
We thus introduce the contrastive object continuity loss as:


\begin{equation}\label{eq:oc_single}
\mathcal{L}^\text{OC}(\textbf{x}_{t}) = \sum_{o^{i}_{t} \in \Omega_{t}} \sum_{o^{j}_{t\!+\!1} \in \Omega_{t\!+\!1}} \lambda_{\text{differ}} \cdot S_C(o^{i}_{t}, o^{j}_{t\!+\!1}),
\end{equation}
where $\lambda_{\text{differ}}$ is defined, using the hyperparameter 
$\beta \in \mathbb{R^{+}}$, as:
\begin{equation}
\label{eq:loc_def}
    \lambda_{\text{differ}} =
    \begin{cases}
        -\beta & \text{if } o_{t+1}^j = \texttt{enc}^*({o}_{t}^i) = \texttt{loc}^*({o}_{t}^i) \\
        1 & \text{else} \\
    \end{cases}
\end{equation}
This loss allows the models to have better internal representation of an object, for which the visual representation can vary across the frames.  


\subsection{General Training Scheme}

The MOC scheme thus overall adds motion and object continuity supervision to a base object detection model, resulting in the final loss function:

\begin{equation}\label{eq:loss_moc}
\small{\mathcal{L}^{\text{MOC}} := \mathcal{L}^{B} + (1 - \lambda_{align}) \cdot \mathcal{L}^\text{M} + \lambda_{align} \cdot \lambda_{\text{OC}} \cdot \mathcal{L}^\text{OC}}.
\end{equation}

\noindent $\mathcal{L}^\text{M}$ represents the batch-wise motion supervision loss and $\mathcal{L}^\text{OC}$ the batch-wise object continuity loss. 
We use $\lambda_{align} \in [0, 1]$, to balance the learning of object detection and object representation. We recommend using a scheduling approach for $\lambda_{align}$, described in the App.~C.4
. An additional $\lambda_{\text{OC}}$ hyperparameter can be used to balance image reconstruction and encoding improvements. 

Concerning our previously suggested error types of object representation learning: the goal of $\mathcal{L}^\text{OC}$ is to improve on the Type II error. As previously identified, improvements on Type II also depend on improvements on the Type I error for which  $\mathcal{L}^\text{M}$ was developed. The ultimate goal of MOC to improve deep object representation learning for downstream reasoning tasks is thus achieved by interdependently improving over both error types.

Lastly, MOC takes advantage of temporal information, while leaving the base architecture unchanged. In this way, it is possible to learn the concept of objects from image sequences, while still allowing for single image inference.


\section{Experimental Evaluations}

Let us now turn to the experimental evaluation of the MOC training scheme. With our experimental evaluations, we wish to investigate whether motion and object continuity biases can benefit object-centric DL for downstream reasoning tasks. We investigate the effect of our MOC scheme on object discovery abilities and on object representation learning, such that these can be used for downstream tasks such as object-centric game playing and few-shot object classification. 
Specifically, we address the following research questions: (\textbf{Q1}) Does MOC benefit object discovery? (\textbf{Q2}) Does MOC improve latent object representations? (\textbf{Q3}) Does MOC improve representations for complex downstream reasoning tasks, such as game playing and few-shot object classification?


\textbf{Experimental setup.} To this end, we consider the SPACE model~\cite{SPACE2020} and a Slot Attention model~\cite{locatello2020slotattention} as base object discovery models, and compare performances of these trained with (+MOC) and without our MOC scheme (baseline, models only optimized for reconstruction). As we are ultimately interested in RL as downstream reasoning task, we train these models on images obtained from different reinforcement learning environments of the Atari 2600 domain \cite{Brockman2016OpenAIG}. As SPACE is the only object discovery model ---to the best of our knowledge--- trained on images obtained from different of the mostly used Atari environments, we focus the bulk of our evaluations on this, but provide results also on a Slot Attention model to show the generality and improvements across different base models. 
Both architectures are visualized in Fig.~7
, with further details and training procedures (\cf App.~B
). 
We specifically focus on a subset of Atari games, namely AirRaid, Boxing, Carnival, MsPacman, Pong, Riverraid, SpaceInvaders and Tennis, many of which were investigated in the original work of SPACE. Note that this subset of Atari games contains a large variance in the complexity of objects, \ie the number, shape, and size of objects, as well as games with both static and dynamic backgrounds, representing a valid test ground. 

Overall, we evaluate $3$ model settings: the original base models, SPACE and Slot Attention (SLAT), and both incorporated in our full MOC training scheme (SPACE+MOC and SLAT+MOC). We further provide specific ablation experiments in which the base model is only trained via $\mathcal{L}^{B}$ and $\mathcal{L}^{M}$, but without $\mathcal{L}^{OC}$. For each experiment, the results are averaged over converged (final) states of multiple seeded runs. We provide details on the hyperparameters in App.~B 
for both models and on the evaluation metrics in each subsection, and further in App.~E
. Unless specified otherwise, the reported final evaluations are performed on a separated, unseen test set for each game. 

As the aim of our work focuses on RL for Atari games, in our experimental evaluations we use a basic optical flow (OF) technique described in App.~C.1 
which provides us with sufficient masks $\hat{\alpha}$ of moving objects. As previously noted, the exact optical flow implementation, however, is not a core component of MOC and can be replaced with any other out-of-the-box OF approach such as \cite{teed2020raft, stone2021smurf}. All figures are best viewed in color.

\textbf{Atari-OCTA.}
As the Atari dataset from \cite{SPACE2020} is not labelled and thus insufficient particularly for evaluating object encodings, we created a novel version of it. For this, we created one labelled dataset per game, where positions, sizes, and classes of all objects are provided. We used OC-Atari \cite{Delfosse2023OCAtariOA} to extract the objects properties. We separate classes of objects that are relevant for successful play from the objects of the HUD (\eg scores, lives), and base our metrics only on this first set, as we investigate MOC in the context of RL as reasoning task. We provide image sequences of $T=4$ for the training data.
For details on how Atari-OCTA (the resulting object-centric and time annotated version of Atari) was created, see App.~A
. Atari-OCTA and the generation code are provided along with this paper. \footnote{\url{https://github.com/k4ntz/MOC}}
Our evaluations were performed on Atari-OCTA.

\begin{figure}[t!]
    \centerline{\includegraphics[width=1.\columnwidth]{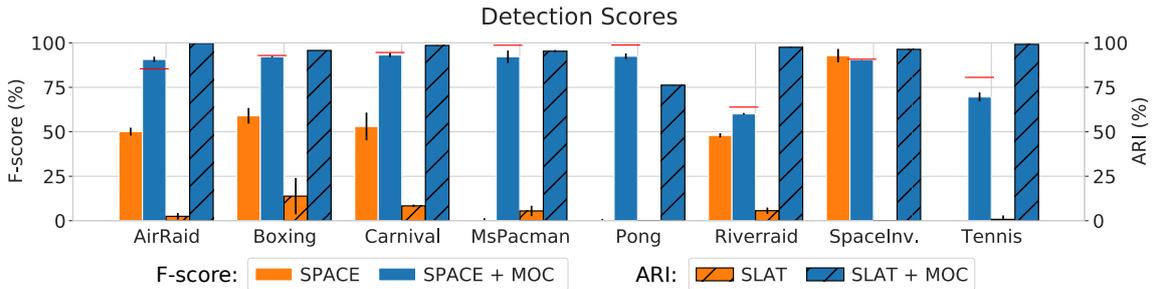}}
    \caption{MOC improves object detection. Final F-scores of SPACE models and Adjusted Random Index of Slot Attention (SLAT), both with and without MOC over frames of different Atari-OCTA games. Training via MOC leads to massive improvements over the set of investigated games. Optical flow F-scores are provided in red. They indicate the potential F-score upper-bound obtainable if using Motion supervision only.}
  \label{fig:f_score} 
\end{figure}
\setlength{\parindent}{11pt}
\textbf{MOC benefits object discovery (Q1).}
Let us now move to the actual research question evaluations.
First, we evaluate the influence of our training scheme on a base model's performances in the originally intended task of object discovery, thus investigating the potential improvements over the Type I error alone. 
For this, we compute the foreground F-scores (further denoted as F-score) of SPACE as in the original work of \cite{SPACE2020} and compare to SPACE+MOC, \ie SPACE incorporated in MOC. As Slot Attention is missing explicit variables for computing the F-scores, we revert to providing foreground Adjusted Rand Index (further denoted as ARI) scores for SLAT as well as SLAT+MOC, as also provided in the original work by Locatello et al.~\cite{locatello2020slotattention}. 

First, Fig.~\ref{fig:f_score} presents the final foreground object detection scores (i.e. F-scores for versions of SPACE and ARI for versions of SLAT) of the different training setups for images from the eight Atari-OCTA games. 
Over the set of these, adding motion and object continuity supervision leads to great improvements in object discovery, with an average improvement from $38\%$ to $85\%$ and $4\%$ to $95\%$ over all games for SPACE+MOC and SLAT+MOC, respectively.

In particular, the base SPACE model on average performs quite unsatisfactorily. The base SLAT model performs even more poorly, importantly indicating SPACE's superiority over SLAT for Atari images, possibly due to the object size prior of SPACE's and its grid-based approach. On games such as MsPacman and as arguably simple as Pong, SPACE and Slot Attention even appear to fail. Here, using our MOC scheme leads to more than $90\%$ increase in object detection final performances.
We also note the reduced performance variance of both +MOC models, suggesting more reliable training via MOC. 

Furthermore, object biases via MOC not only improve final object detection performance, but also aid in obtaining such performances with fewer training iterations. Tab.~\ref{tab:fsco_steps} presents the average number of steps required for SPACE to obtain $50 \%$ of the validation F-scores of the different training schemes on the $8$ Atari games. Over all games SPACE+MOC models approach convergence in much fewer number of steps. In fact, SPACE is able to reach an F-score above $50\%$ on only $5$ out of $8$ games. For these games, it takes on average $1820$ steps to reach this threshold, compared to $280$ on average for SPACE+MOC on all games, hence leading to more than $7$ times faster learning. 

We provide detailed F-score progressions on all games, as well as final precision vs recall curves in App. (\cf Fig.~9 
\& 10
). We also provide results with SPACE+MOC w/o OC indicating that updating object encodings via the OC loss in MOC on average shows equivalent object detection performances as without. This is an intuitive finding and shows that the motion loss, intentionally developed for tackling the Type I error, achieves its purpose.
Our results in summary indicate a strong benefit of object location biases via MOC for object detection, thus affirming \textbf{Q1}. 

\begin{table}[t!]
\caption{Faster convergence through MOC. The average number of steps needed to get an F-score $> 0.5$ ($\infty$ if never reached) on the test set for SPACE, with and without MOC on Atari-OCTA. Lower values are better. For each model, we also provide the average. A complete evolution of the F-scores is given in App.~D
.}
\label{tab:fsco_steps}
\centering
\setlength{\tabcolsep}{3pt}
\begin{tabular}{@{}lcccccccc||c@{}}
\toprule
Game             & {Airraid}      & {Boxing}         & {Canival}       & {MsPac.}       & {Pong}            & {Riverraid}      & {Sp.Inv.}      & {Tennis}         &  {Avg}           \\ \midrule
SPACE            & 2600           & 2200         & 420          & $\infty$       & $\infty$        & 3400           & 430          & $\infty$       & 1800         \\ 
SP.+MOC & \textbf{300} & \textbf{240} & \textbf{350} & \textbf{350}   & \textbf{270}   & \textbf{270} & \textbf{290} & \textbf{260}   & \textbf{280} \\ \bottomrule
\end{tabular}
\end{table}

\textbf{MOC improves latent object encodings (Q2).}
In the previous section, we presented improvements of base object discovery models, SPACE and SLAT, via MOC, focusing on the task of object discovery for which these models were mainly developed for and evaluated on. However, our investigations in Fig.~\ref{fig:motivation} exhibited that even when the object detection performance was promising (\eg Space Invaders) the object representations themselves proved less useful for a downstream reasoning task as simple as object classification. Thus, apart from improvements in the detection itself, additional improvements need to be achieved in terms of optimal object representations for such models to actually be integrated into a more complex downstream setup. With \textbf{Q2}, we thus wish to investigate the effect of the MOC scheme on improving a base model's Type II errors, \ie the usefulness of a model's latent object representations. 

\begin{figure}[b]
    \centering
    \includegraphics[width=1.\columnwidth]{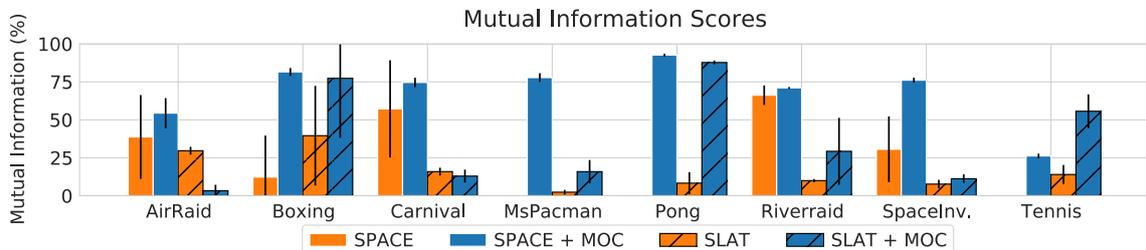}
    \caption{\textbf{MOC leads to more optimal object encodings as indicated via mutual information score.} The adjusted mutual information of object encodings from SPACE and Slot Attention (SLAT), both with and without MOC, of Atari-OCTA are presented (mean $\pm$ std). Higher average values are better.}
    \label{fig:AMI} 
\end{figure}

To answer \textbf{Q2}, we focus on the adjusted mutual information (AMI) of the encodings. Essentially, one computes a clustering on the latent object encodings and compares the clustering to a ground truth clustering (\cf App.~F 
for details). Fig.~\ref{fig:AMI} presents the AMI for the two configurations: base model (i.e. SPACE and SLAT) and +MOC for training with MOC. One can observe immense improvements for MOC trained models in comparison to the base models. Averaged over all games we observe an increase in AMI from $26\%$ to $69\%$ and $16\%$ to $37\%$ for SPACE+MOC and SLAT+MOC, respectively.

In addition, one can observe the benefits of leveraging both $\mathcal{L}^\text{OC}$ \textit{and} $\mathcal{L}^\text{M}$ in an ablation experiment for SPACE in which SPACE+MOC w/o OC is trained only via the motion supervision loss. Results can be found in App.~D.5
. As the object encodings in the original SPACE and in SPACE+MOC w/o OC are only optimized for reconstruction, there is little supervision to produce distinct object representations. Our results thus indicate that $\mathcal{L}^\text{OC}$ provides a strong supervisory signal on the latent encoding space. Lastly, as seen by the reduced cross validation variance, on average MOC produces less sensitive models (supporting the findings of Fig.~\ref{fig:f_score}). We provide detailed AMI scores in App. (\cf Fig.~13
).

These results overall highlight the improvements of MOC, particularly the benefits of integrating both supervisory mechanisms into one training scheme, as well as $\mathcal{L}^\text{OC}$'s benefit for improving on Type II errors.
Conclusively, we can affirm \textbf{Q2}: MOC does improve latent encodings. 

\textbf{MOC improves object encodings for downstream reasoning tasks (Q3).}
AMI is one way of measuring the quality of latent encodings. Ultimately, we are interested in integrating unsupervised object discovery models into more complex downstream tasks, \eg integrating these into RL. In the following, we focus on evaluating object encodings for two specific downstream reasoning tasks, namely few-shot object classification and Atari game playing. The resulting performances in these downstream tasks thus act as important additional evaluations on the quality of learned object encodings and act as the main motivation behind our MOC scheme.

Let us begin with few-shot object classification. 
For each game, we optimize linear ridge regression models on $1$, $4$, $16$ and $64$ object encodings of each object class and measure the classification accuracy on a separate held-out test set. We thus assume useful encodings to correspond to encodings that are similar for objects of the same class. As such, these representations should be easier to be differentiated by a simple (linear) classification method. 

\begin{figure}[t!]
    \centering
    \includegraphics[width=.75\columnwidth]{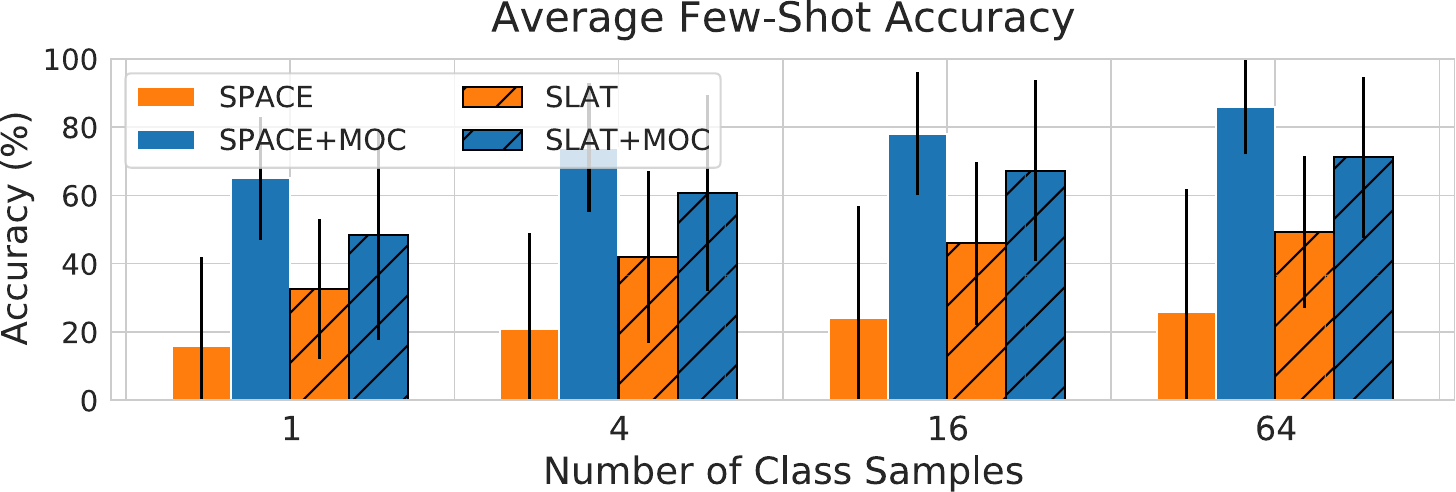}
    \caption{MOC improves object representations for the few-shot classification task. Average few-shot accuracy performance (in $\%$) based on latent object representations. We provide a ridge regression model with $1$, $4$, $16$ and $64$ encodings per class and test it on a held-out test set. The values are averaged over the $8$ investigated Atari-OCTA games for SPACE and Slot Attention (SLAT), with and without MOC. Detailed, per game results are in App.~G.2
    .}
    \label{fig:evo_fsa}
\end{figure}

The results can be seen in Fig.~\ref{fig:evo_fsa}.
Specifically, we see a large boost in classification performance via the full MOC training scheme, visible both in SPACE+MOC and SLAT+MOC performances. Note the strong contrast, particularly for SPACE in test accuracy in the very small data regime, \eg when the classifier has only seen 1 object encoding per class. Here, SPACE+MOC reaches $65\%$ average accuracy, $3.5 \times$ higher than SPACE and $1.3 \times$ higher than SPACE+MOC w/o OC (\cf App.~D.3
). 
Detailed results on each game are provided in App.~(\cf Fig.~12 
and additional qualitative results can be found in App.~D.6 
and D.7
.

We finally evaluate MOC in the context of RL for game playing as a representative complex downstream reasoning task and investigate in how far MOC can improve a base model's performance such that its learned representations can be used in playing Atari games, as done by \cite{Delfosse2023InterpretableAE} in logic oriented settings. We base our evaluations on concept-bottleneck \cite{KohNTMPKL20conceptbottleneck} RL agents that are initially trained using perfect information (Info*) with the Dueling Q-learning algorithm \cite{WangSHHLF16duelingdqn} and focus here on playing a subset of the games from our previous evaluations (Boxing and Pong). These games are the only two from Atari-OCTA with a fixed number of objects, which allows for a very simple policy architecture, and for which perfect information obtained from the RAM using \cite{anand2019atariari} is available. We note that although many RL agents might possibly provide higher overall game scores than the ones we focus on, the point to make here is that training object encodings via MOC can greatly improve \textit{a} baseline (object-centric) RL agent. 

For these experiments, we focus only on SPACE as base model (rather than SLAT) due to its superior base performance on object detection in Atari images. We next replace the object extraction from RAM with object extraction performed by SPACE, and SPACE+MOC models but keep the initially learned policies. For this, we transform the \texttt{loc} variables of these models into $(x, y)$ coordinates (ignore sizes). For object classes, we use the previously evaluated 
few-shot classification models on the object encodings (with $16$ samples per class, \cf Fig.~12
). The object coordinates are given to a $3$ fully connected layers network, which predicts the Q-values. We provide additional comparisons to human-level performance (H) and random agents (R). 

As can be seen in Tab.~\ref{tab:rl}, MOC greatly empowers the evaluated object-centric agents. On average, SPACE+MOC-based agents largely outperform SPACE-based agents, and even obtain higher average scores than humans on Boxing. For both games, the benefits of our approach is even more remarkable on mean scores obtained by best playing agents, due to the fact that a slightly more accurate object detector results in more reliable states, and thus a better performing agent.
Our experimental evaluations allow us to validate that MOC benefits object-centric reasoning tasks spanning from RL agents game playing capabilities to few-shot object classification (\textbf{Q3}).

\begin{table}[t!]
\caption{MOC (via SPOC) allows object-centric playing via concept-bottleneck models. Average scores among all agents (avg) and of the best agent (best) of object-centric agents that detect objects based on SPACE and SPOC. Scores for each seed are in App.~G.3.
Additional base comparisons are agents with perfect information (Info*), random agents and human (from \cite{vanHasseltGS16ddqn}).}
\label{tab:rl}

\centering
\setlength{\tabcolsep}{5pt}
\resizebox{0.7\columnwidth}{!}{
\begin{tabular}{@{}l|l|cc|ccc@{}}
\toprule
                            & Method & SPACE                                                           & SP.+MOC           
                            & Info*                      & Random  & Human  \\ \midrule
\multirow{2}{*}{\rotatebox[origin=c]{90}{avg}} 
& Boxing & \text{-}$3.5\mbox{\scriptsize$\pm 4.3$}$ & $\textbf{8.4}\mbox{\scriptsize $\pm9.9$ } $ & $36\mbox{\scriptsize$\pm 17$ }$ & $\text{-}0.5\mbox{\scriptsize$\pm2. $ } $ & $4.3$ \\
& Pong   & \text{-}$21\mbox{\scriptsize$\pm 0.5$ }$ & \text{-}$\textbf{11}\mbox{\scriptsize$\pm 12$}$ & $20\mbox{\scriptsize$\pm1.3$}$ & \text{-}$21\mbox{\scriptsize$\pm0.3$} $ & $9.3$ \\ 
\midrule
\multirow{2}{*}{\rotatebox[origin=c]{90}{best}} 
& Boxing & $3.8\mbox{\scriptsize$\pm 6.5$} $ &  $\mathbf{22}\mbox{\scriptsize$\pm 14$} $ & $52\mbox{\scriptsize$\pm3.5$ }$ & $2.6\mbox{\scriptsize$\pm3.3$}$ & - \\
& Pong   & \text{-}$20\mbox{\scriptsize$\pm0.8$ }$ & $\textbf{4.8}\mbox{\scriptsize$\pm 11$}$ & $21\mbox{\scriptsize$\pm 0.$}$ & \text{-}$20\mbox{\scriptsize$\pm 0.7$}$& - \\
\bottomrule
\end{tabular}}
\end{table}

\section{Related Work}

Our work touches upon several topics of ML research. In the following, we discuss related works from unsupervised object detection, and object tracking and self-supervised learning.


\textbf{Unsupervised Object Detection.}
Object-centric DL has recently brought forth several exciting avenues of unsupervised and self-supervised object discovery research by introducing inductive biases to neural networks to extract objects from visual scenes in an unsupervised manner \cite{eslami2016air, crawford2020exploiting, kosiorek2018sequential, jiang2019scalor, stelzner2019faster, ZhangHP19, burgess2019monet, EngelckeKJP20, GreffKKWBZMBL19, SPACE2020, locatello2020slotattention, JiangA20, smirnov2020marionette}. We refer to \cite{greff2020binding} for a detailed overview. 
However, until recently \cite{seitzer2022bridging, SinghDA22} these approaches have shown limited performances on more complex images. All of these mentioned works focus on object discovery from independent images. 
Importantly, although most of these approaches are motivated with the benefits of object representations for more complex downstream tasks, none of these apply additional constraints onto the latent object representations and only optimize object encodings for reconstruction. 
Lastly, among these works are more recent approaches that could provide improved baseline performances than SPACE and the vanilla Slot Attention model (\eg \cite{seitzer2022bridging, SinghDA22}), however MOC can also be applied to these and these should be seen as orthogonal to our work.

\textbf{Unsupervised Object Detection from Videos.}
Leveraging the supervisory signal within video data for object detection goes back several years \eg to approaches based on conditional random fields \cite{SchulterLRB13}. More recently, DL approaches, \eg \cite{zhao21} explicitly factorize images into background, foreground and segmentation masks. 
Also, \cite{DuSUT021} argue for the value of motion cues and showcasing this on physical reasoning tasks via a dynamics-based generative approach. 
\cite{zhou2021target} do multi-object segmentation via a carefully designed architecture of multiple submodules that handle certain aspects of the overall task.
This idea is also taken up by \cite{tangemann2021unsupervised} who introduce an approach for generative object-centric learning via the property of \textit{common fate}.
The recent works SAVi \cite{kipf2021SaVi} and SAVI++ \cite{elsayed2022savi++} improve the object discovery abilities of Slot Attention by using optical flow and depth signals, respectively, as a training target. 
Where many of these works present specifically designed architectures for video processing, MOC represents a model-agnostic scheme which, in principle, can incorporate any base object detection model and in this sense also be applicable to such tasks as semantic world modeling and object anchoring~\cite{PerssonMRL20}. Additionally, although many mentioned works identify the importance and benefits of motion for object discovery, only \cite{DuSUT021} also focus on the advantages of learning good object representations for downstream tasks. 

\textbf{Optical Flow.}
Optical flow methods aim at finding translation vectors for each pixel in an input image to most accurately warp the current frame to the next frame. In other words, optical flow techniques try to find pixel groups with a common fate. 
Estimating optical flow is a fundamental problem of computer vision, starting with such approaches as the Gunnar-Farneback Algorithm \cite{farneback2003two} (for other traditional approaches, please refer to \cite{fleet2006optical}). By now it is widely used in a variety of applications such as action recognition \cite{cai2019temporal, lee2018motion}, object tracking \cite{kale2015moving, zhou2018deeptam}, video segmentation \cite{liu2020efficient, li2021dynamic}. Most recent works are based on DL approaches \cite{dosovitskiy2015flownet, ilg2017flownet, luo2021upflow, hur2020optical, teed2020raft, stone2021smurf, jeong2022imposing}. Recently optical flow has also been applied to RL settings  \cite{goel2018unsupervised, yuezhang2018initial} which hint the downstream RL agent at possibly important object locations, but do not perform object representation learning. As previously mentioned, optical flow is considered as a preprocessing step in MOC, thus any suitable OF method can in principle be used for a specific MOC implementation.

\textbf{Self-Supervised Learning.}
The idea of motion supervised learning is related to the large field of self-supervised learning. 
In fact the motion supervision aspect of MOC stands in line with other approaches \cite{kipf2021SaVi, elsayed2022savi++, yang2021self} and notably \cite{Bao2022discovering} and confirms their findings on the general benefit of motion priors for object detection, however these works do not specifically improve or otherwise evaluate obtained object encodings for anything other than object detection.
Aside from this, the $\mathcal{L}^{\text{OC}}$ loss of MOC can be considered as a form of contrastive loss, popularly used in self-supervised learning. Among other things, recent works apply patch-based contrasting \cite{henaff2020data}, augmentations \cite{chen2020simple}, cropping \cite{he2020momentum} or RL based contrasting \cite{laskin2020curl}. In comparison, MOC contrasts encodings between consecutive time steps.
These approaches do not perform object representation learning, but rather try to improve the full encoding of an image.


\section{Limitations and Future Work}
As we focus evaluations on Atari environments for classification and game playing tasks, we make use of a very simple optical flow technique. A more advanced optical flow approach is necessary for more complex environments. 
The application of MOC to other types of object discovery models is a necessary next step in future work. Additionally, the $\mathcal{L}^{\text{OC}}$ of our SLAT+MOC implementation produces larger variances in AMI than that of SPACE+MOC (\cf Fig.~\ref{fig:AMI}). Future work should investigate ways of stabilizing this.
MOC contains a bias towards moving objects, however it is desirable for a model to also detect stationary objects. Although the scheduling approach tackles this issue, additional investigations are necessary. 
The bottleneck-based \cite{KohNTMPKL20conceptbottleneck} RL agents of \textbf{Q3} are somewhat limited, as additional techniques to increase detection robustness (\eg Kalman filters) were not used. 
Certainly, more complicated models and RL algorithms would be worth investigating in future research, such as~\cite{Delfosse2024InterpretableCB}. 
Lastly, certain environments also include static objects that are important to interact with, we would thus like to investigate object recognition based on agent-environment interactions. Following this idea, we think that performing representation learning based on the RL signal (\ie cumulative reward) is another interesting line of research.

\section{Conclusions}
Already young children quickly develop a notion of object continuity: they know that objects do not pop in and out of existence. Here, we have demonstrated that deep networks also benefit from this knowledge. Specifically, 
we have shown that properties of objects that are observed from time can be leveraged to improve object representation learning for downstream tasks. This MOC training scheme, consisting of a motion supervision and object continuity contrastive loss, greatly improves a base model's object discovery and representation performances. In this way MOC collectively tackles two interdependent error sources of current object discovery models. 
Finally, along with this paper, we provide the novel dataset Atari-OCTA, an important step for evaluating object discovery in complex reasoning tasks such as game playing.

Apart from performance benefits that object-centric approaches can bring to deep learning, they particularly play an important role in human-centric AI, as such machines can perform inference and provide explanations on the same object-level as human-human communication \cite{kambhampati2022symbols}. Additional avenues for future work include investigating interactive object learning, such as agent-environment interactions, but also human-machine interactions for more reliable and trustworthy AI models.

\endgroup  

\section*{Ethical Statement}

Our work aims to improve the object representations of object discovery models, specifically targeting the improvements of their use in additional modules in downstream reasoning tasks. With the improvements of our training scheme, it is feasible to integrate the findings of unsupervised object discovery methods into practical use-cases. A main motivation, as stated in our introduction, is that such an integration of high-quality object-centric representations is beneficial for more human-centric AI. Arguably, it seems beneficial for humans to perceive, communicate and explain the world on the level of objects. Integrating such level of abstraction and representation to AI agents is a necessary step for fruitful and reliable human-AI interactions. 

Obviously, our work is not unaffected from the dual-use dilemma of foundational (AI) research. And a watchful eye should be kept, particularly on object detection research which can easily be misused, \eg for involuntary human surveillance. However, our work do not, to the best of our knowledge, pose an obvious direct threat to any individuals or society in general.

\subsubsection*{Acknowledgements}
The authors thank the anonymous reviewers of ECML 2023 for their valuable feedback. This research work has been funded by the German Federal Ministry of Education and Research and the Hessian Ministry of Higher Education, Research, Science and the
Arts (HMWK) within their joint support of the National Research Center for Applied Cybersecurity
ATHENE, via the ``SenPai: XReLeaS'' project. It also benefited from the HMWK cluster projects ``The Third Wave of AI'' and ``The Adaptive Mind'' as well as the Hessian research priority program LOEWE within the project ``WhiteBox''.

\newpage

{\small
\bibliographystyle{ieee_fullname}
\bibliography{main}
}

\onecolumn
\begin{center}
\textbf{\large Supplemental Materials}
\end{center}
\setcounter{section}{0}
\renewcommand{\thesection}{\Alph{section}}

In the following, you can find details on how our Atari-OC dataset was generated (\ref{app:dataset_details}), details and mathematical formalism for the original SPACE and Slot Attention baseline models (\ref{appendix:architecture_details}), further details on the computations of our MOC training scheme (\ref{appendix:moc_details}), additional visual results (\ref{appendix:visual_results}), the hyperparameters used for this baseline and for our MOC scheme (\ref{appendix:hyperparameters}), details on the evaluation metrics we used (\ref{appendix:evaluation_metrics}), all our results in table to allow future comparisons (\ref{appendix:numerical_results}), and finally per games details on the Atari-OC dataset generation (\ref{appendix:ocatari_detail}).

\section{The Atari-OC dataset}
\label{app:dataset_details}

\begin{wraptable}{r}{0.35\textwidth}
\vspace{-1cm}
\centering
\captionsetup{belowskip=-1em}
\begin{tabular}{@{}ll@{}}
\toprule
Game           & Agent Type \\ \midrule
Air Raid       & DQN        \\
Boxing         & Random     \\
Carnival       & Rational DQN  \\
Ms. Pacman      & Rational DQN  \\
Pong           & Rational DQN   \\
Riverraid      & DQN        \\
Space Invaders & Rational DQN   \\
Tennis         & Random     \\ \bottomrule
\end{tabular}
\caption{Agents used for Dataset Creation. DQN correspond to DQN agents from, Rational DQN perform better, and thus explore more changing environments \cite{Delfosse2021AdaptiveRA}}
\label{table:agentusage}
\end{wraptable}
In this work we focus on sequence lengths of 4 frames and use Open AI Gym \cite{brockman2016openai} to generate the training data based on Atari environments. 
In particular, we use a trained agent for gathering the dataset to make sure diverse scenes are captured. For practical reasons we used a rational agent (following \cite{Delfosse2021AdaptiveRA}) if available, simply selected a random agent or mainly because of Riverraid trained a DQN-based Agent. 
In the later case, we select a sufficient agent trained with only 5,000,000 samples following the implementation of DQN provided by Mushroom-RL \cite{deramo2020mushroomrl}\footnote{The code can be found here: \url{https://github.com/MushroomRL/mushroom-rl/blob/dev/examples/atari_dqn.py}, Accessed 2022/03/10}. Table~\ref{table:agentusage} provides an overview for each game.
Finally, we sample between every 20-100 frames (depending on the game) and in this way collect $2^{13} = 8192$ frame-stacks (each with 4 images) for each game as the training set and 1024 stacks for validation and test sets. In comparison, \cite{SPACE2020} were originally using 50000 and 5000 images, respectively.

\subsection{Groundtruth Object Labeling}

Quantitatively evaluating a model's performance for object detection on Atari environment images is non-trivial, as no groundtruth object information is provided. The authors of SPACE \cite{SPACE2020}, for example, only provided qualitative reconstruction results to indicate SPACE's performance. 
To circumvent this issue we used two approaches for finding bounding boxes and object class labels of the Atari images. The first was to use the RAM hacking approach of AtariARI \cite{anand2019atariari}. However, this only covers a subset of games and at times provides incomplete information e.g. only positions with unknown offsets or missing positional information of stationary objects like the score or extra lives. 

The second, more heuristic approach was thus to find contours of object colors in an image using OpenCV \cite{opencv_library}. This required several manual steps in collecting the individual RGB values of the objects' colors. If a color was used for various object classes we applied further filtering based on additional properties such as position and size of the bounding box. 
All of those steps for each individual game are listed in section~\ref{appendix:ocatari_detail}. 

\begin{table}[ht]
\centering
\captionsetup{justification=justified}
\begin{tabular}{@{}ll@{}}
\toprule
\textbf{Game} & \textbf{Condition} \\ 
\midrule
Air Raid & $0.063 < y$ \\ 
Boxing & $0.148 < y < 0.859$ \\ 
Carnival & $0.117 < y$  \\ 
MsPacman & $y < 0.813$ \\ 
Pong & $0.164 < y$  \\ 
Riverraid & $y < 0.766$ \\ 
SpaceInvaders & $[\text{Mystery Ship present}] \lor 0.125 < y$  \\ 
Tennis & $(0.063 < y \land y_{max} < 0.469) \lor (0.531 < y \land y_{max} < 0.906)$ \\ 
\bottomrule
\end{tabular}
\vspace{0.3cm}
\caption{A bounding box $b:=[x, y, x_{max}, y_{max}]^T \in [0,1]^4$ is considered \textit{relevant} if \textbf{Condition} holds. This simply restricts objects to the game board and filters out the scoreboard reasonably well (for Riverraid the fuel gauge is cut off unintentionally, which introduces a minor error).}
\label{table:filteringrelevantobjects}
\end{table}
Finally, we restricted ground truth detected objects to focus on relevant objects, i.e. that lay in the playing field of each Atari environment, motivated by the application to an RL downstream task. For this we provide simple bounding box constraints per game as seen in Table~\ref{table:filteringrelevantobjects}.

\section{Details on the original baselines model}
\label{appendix:architecture_details}
In this section, we present in more details the two baselines model: SPACE and Slot Attention. We also provide a visual description of both models hereafter.

\begin{figure}[ht!]
    \centering
    \includegraphics[width=1.\textwidth]{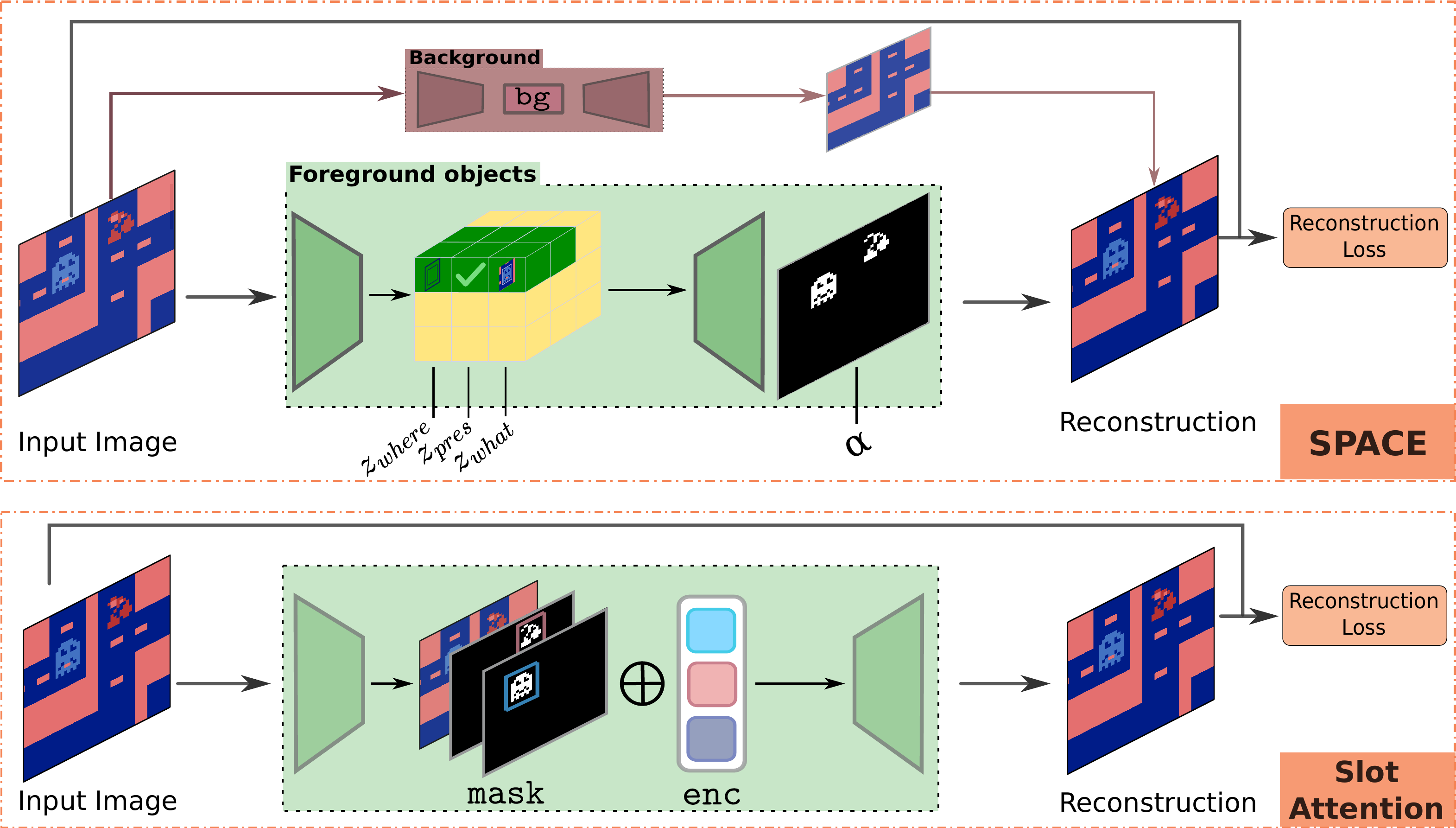}
\caption{A visual description of both used baseline models: SPACE and Slot Attention. 
In SPACE, the model separates the background from the foreground. For the foreground components (\ie objects), the image divided in a grid. For each cell, SPACE encodes the presence of an object in $z_{pres}$, the potential object's location (inside the cell) in \texttt{loc} and its representation in \texttt{enc}. Here, the general \texttt{loc} variable is actually encoded through the grid definition and both the $z_{pres}$ and \texttt{loc} variables.
Slot Attention encodes both the background and the objects in slots ($3$ in the depicted example). These slots are composed of \texttt{mask} variables, (from which \texttt{loc} can be recovered), and \texttt{enc} variables, which represent the location and encodings of the objects.
Figure best viewed in color.}
    \label{fig:models_representations} 
\end{figure}

\subsection{SPACE}
\label{appendix:space}
SPACE uses two variational autoencoders (VAE) to separately learn latent representations of the foreground components (\ie objects), denoted $\textbf{z}^{fg}$, and the background, $\textbf{z}^{bg}$.
The background consists of $K$ background segments $\{\textbf{z}_k^{bg}\}_{k=1}^K$, while the foreground is divided into a $H \times W$ grid, with individual foreground components $\{\textbf{z}_i^{fg}\}_{i=1}^{H \times W}$, allowing for the detection of $c$ ($\leq H \times W$) objects.
A specific scene $\textbf{x}$ is thus modeled as
\begin{equation}
\label{eq:space_reconstruction}
\thickmuskip=2mu
p(\mathbf{x} \mid \mathbf{z}^{\mathrm{fg}}, \mathbf{z}^{\mathrm{bg}}) = \alpha p(\mathbf{x} \mid \mathbf{z}^{\mathrm{fg}}) + (1-\alpha) \sum_{k=1}^{K} \pi_{k}p(\mathbf{x} \mid \mathbf{z}_{k}^{\mathrm{bg}}),
\end{equation}
where $\alpha = f_\alpha(\textbf{z}^{fg})$ represents the foreground mixing probability and $\pi_{k}\!=\!f_{\pi_{k}}(\textbf{z}^{bg}_{1:k})$ the mixing weight assigned to the background subdivided among K background components.

Each foreground cell is modelled with latent variables $(\textbf{z}_i^{pres}, \textbf{z}_i^{where}, \textbf{z}_i^{what})$, where $\textbf{z}_i^{pres}$ represents the probability of an object being present in the scene ($z_{pres}$), $\textbf{z}_i^{where}$ (jointly these describe the $\texttt{loc}$ information), relative to the grid cell, and the size of a potential object and $\textbf{z}_i^{what}$ encodes the object's information encoding ($\texttt{enc}$).
In case of environments with overlapping objects, an additional $\textbf{z}_i^{depth}$ variable can be used. 
Overall the foreground prior is given as:

\begin{equation}
    p(\mathbf{z}^{\mathrm{fg}})=\prod_{i=1}^{H \times W} p(\mathbf{z}_{i}^{\text {pres }}) \left( p(\mathbf{z}_{i}^{\text {where }}) p(\mathbf{z}_{i}^{\text {what }}) \right) ^{\mathbf{z}_{i}^{\text {pres }}}.
\end{equation}
The background prior is given as 
\begin{equation}
    p(\mathbf{z}^{\mathrm{bg}})=\prod_{i=1}^{K} p(\mathbf{z}_{k}^{c} | \mathbf{z}_{k}^{m}) p(\mathbf{z}_{k}^{m} | \mathbf{z}_{<k}^{m}).
\end{equation}
Here, similar to GENESIS~\cite{engelcke2020genesis}, $\mathbf{z}_{k}^{m}$ models the mixing probabilities $\pi_k$ of the components and $\mathbf{z}_{k}^{c}$ models the RGB distribution $p(\mathbf{x}|\mathbf{z}^\mathrm{bg}_k)$ of the $\text{k}^\text{th}$ background component as a Gaussian $\mathcal{N}(\mu^{bg}_i, \sigma^2_{bg} )$.
Finally, the following ELBO:
\begin{equation}
\thickmuskip=0mu
\mathcal{L}^{\text{SPACE}}(\mathbf{x})=
\mathbb{E}_{q(\mathbf{z}^{\mathrm{fg}}, \mathbf{z}^{\mathrm{bg}} \mid \mathbf{x})}[\mathcal{L}_{dec}(\mathbf{x}) + \mathcal{L}^{\mathrm{bg}}_{enc}(\mathbf{x}) + \mathcal{L}^{\mathrm{fg}}_{enc}(\mathbf{x})],
\end{equation}
is used to train the model, with: 
\begin{align}
\label{eq:loss_dec}
\mathcal{L}_{dec}(\mathbf{x}) &= \log p(\mathbf{x} \mid \mathbf{z}^{\mathrm{fg}}, \mathbf{z}^{\mathrm{bg}}) \\
\mathcal{L}^{\mathrm{bg}}_{enc}(\mathbf{x}) &=-\!\sum_{k=1}^{K} D_{\mathrm{KL}}(q(\mathbf{z}_{k}^{\mathrm{bg}} \mid \mathbf{z}_{<k}^{\mathrm{bg}}, \mathbf{x})
    p(\mathbf{z}_{k}^{\mathrm{bg}} \mid \mathbf{z}_{<k}^{\mathrm{bg}}))\\
\mathcal{L}^{\mathrm{fg}}_{enc}(\mathbf{x}) &=-\!\sum_{i=1}^{H \times W} D_{\mathrm{KL}}(q(\mathbf{z}_{i}^{\mathrm{fg}} \mid \mathbf{x}) p(\mathbf{z}_{i}^{\mathrm{fg}}))]
\end{align}

We note that object discovery models such as SPACE~\cite{SPACE2020} heavily rely on priors for object detection and on the accurate modelling capacities of both VAEs. Too much modelling capacity of either the foreground or background VAE can lead to one learning the whole decomposition-recomposition of the scene, leading to a high hyperparameter sensitivity. This is illustrated by the fact that SPACE's authors successfully trained one model to jointly encode and decode images from $10$ different games, among which MsPacman, but could not find a correct architecture and hyperparameters setup to train on these games individually. This explains why SPACE original authors report qualitative results on MsPacman, obtained from a jointly trained model.

\subsection{Slot Attention}
\label{appendix:slot}
Slot Attention models do not separate the foreground objects from the background, but encode the background as one object. Specifically, these models encode objects (and the background) in $L$ slots $\textbf{z} = \{\textbf{z}_l\}_{l=1}^{D \times L}$, which are composed of encodings $\{\textbf{z}^{enc}_l\}_{l=1}^{D \times L}$ corresponding to the \texttt{enc} variables of Sec.~\ref{sec:methods} and obtained via projected filtering masks $\{\alpha_l\}_{l=1}^{N \times L}$ corresponding to the \texttt{loc} variables of Sec.~\ref{sec:methods}. 

A visual scene $\textbf{x}$ is thus modeled as:
\begin{equation}
\label{eq:slat_reconstruction}
\thickmuskip=2mu
p(\mathbf{x} \mid \mathbf{z}) = \sum_{l=1}^{L} \alpha_l \cdot p(\mathbf{x} \mid \mathbf{z}_l^{enc}).
\end{equation}

We refer to the original work for further details~\cite{locatello2020slotattention}.

\section{Details on MOC computations}
\label{appendix:moc_details}

\subsection{Optical flow implementation}
\label{appendix:simple_optical_flow}
\begin{figure}[!ht]
    \centerline{\includegraphics[width=0.6\linewidth]{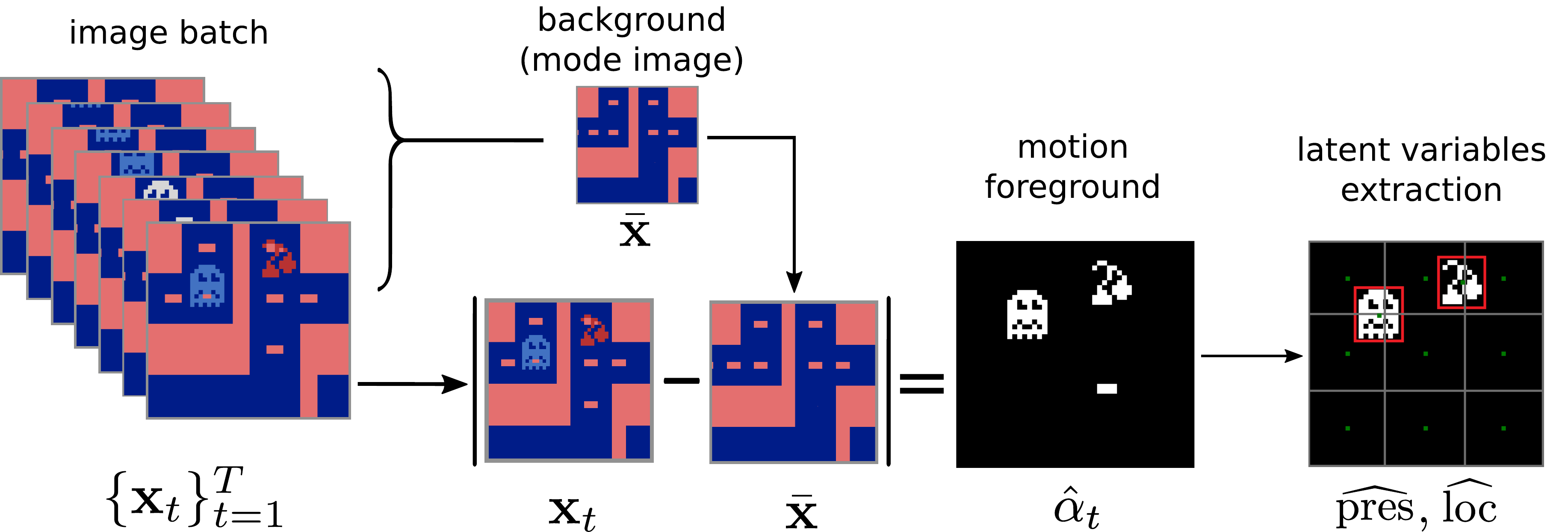}}
    \caption{Motion information (with a simple optical flow technique) allows to directly detect potential objects. The background (\ie mode image) is used to deduce objects in a specific frame.}
  \label{fig:motion_extraction} 
\end{figure}

We here showcase the simple optical flow approach that we use in our experimental evaluations with Atari-OC. While many optical flow methods have been developed to handle complicated scenes, they are not necessary for our Atari dataset. Thus we both explain optical flow intuition and examplify it in Fig.~\ref{fig:motion_extraction}.
Our simple optical flow approach consists in comparing an initially computed background ($\Bar{\textbf{x}}$) to each frame. For static environments, $\Bar{\textbf{x}}$ can be computed using the mode (\ie most common) value of every pixel from a batch of pre-selected images.
We then consider the difference between a frame $\textbf{x}_t$ and $\Bar{\textbf{x}}$ and apply a threshold to obtain the binary mask $\hat{\alpha} = \mathds{1}_{\textbf{x}_t - \Bar{\textbf{x}}>\eta} $, as a hint towards a potential object (\cf Fig.~\ref{fig:motion_extraction}). Let us denote, for every variable $v$ of a model, $\hat{v}$ as the corresponding variable obtained through optical flow estimation (as in Fig.~\ref{fig:time_supervision}).
From $\hat{\alpha}$ we perform simple image processing techniques to identify contours and, consecutively, a set of $\hat{c}$ bounding boxes, from which the $\widehat{z_{pres}}$ and $\widehat{\texttt{loc}}$ variables are obtained. 
If multiple objects match in the same grid cell, the largest bounding box is selected.

We here also provide in Tab.~\ref{table:of_fscores} the F-scores of this simple optical flow method on the human-anotated evaluation sets of every environment of our dataset. The scores obtained by SPACE + MOC on AirRaid and Riveraid are acutally above the ones of the optical flow baseline, showing that the generalisation capacity of the neural network can compensate for non-perfect guidance method.

\begin{table}[!ht]
\centering
\setlength{\tabcolsep}{4.5pt}
\begin{tabular}{@{}l | cccccccc@{}}
F-sc. & Airraid  & Boxing  & Carnival  & MsPacman  & Pong  & Riverraid  & SpaceIn.  & Tennis  \\ \midrule
OF & 85.4 & 92.9 & 94.6 & 98.7 & 98.8 & 63.9 & 90.8 & 80.6 
\end{tabular}
\caption{F-scores of our baseline Optical-Flow method on the different Atari-OC games. Using a different optical flow implementation could help to obtain better scores here, leading to better MOC supervision.}
\label{table:of_fscores}
\end{table}

\subsection{MOC Details for SPACE}
\label{appendix:spoc}
Within object discovery methods such as \cite{SPACE2020} object representations consist of $3$ variables: $\texttt{enc}$ (for content encoding), $\texttt{loc}$ (location and size), and $z_{pres}$ a variable that indicates the presence of an object relative to a predefined grid. The model also learns to predict a foreground mixing probability, $\alpha$, based on $\texttt{fg}$. The input image, $\textbf{x}$, is then reconstructed following:
\begin{equation}
\label{eq:reconstruction}
\thickmuskip=2mu
p(\mathbf{x} \mid \texttt{fg}, \texttt{bg}) = \alpha p(\mathbf{x} \mid \texttt{fg}) + (1-\alpha) p(\mathbf{x} \mid \texttt{bg}).
\end{equation}
The overall loss to train the model is a combination of losses on the foreground encoding, the background encoding and the reconstructed image. More specific details on, \eg SPACE can be found in the App.~\ref{appendix:space}.

For each frame $\textbf{x}_t$, made of $h \times w$ pixels ($pix$), we obtain $\hat{\alpha}$, $\widehat{z_{pres}}$ and $\widehat{z_{where}}$ as described in App.~\ref{appendix:simple_optical_flow} and compute: 
\begin{align}
\mathcal{L}_\alpha(\textbf{x}_t)& := \sum_{pix \in \textbf{x}_t} || \alpha_{pix} - \hat{\alpha}_{pix}||^2\notag\\
\mathcal{L}_{pres}(\textbf{x}_t) & := \sum_{i=1}^{N} ||z_{pres_{i,t}} -  \widehat{z_{pres_{i, t}}}||^2 \\
\mathcal{L}_{where}(\textbf{x}_t) & :=  \sum_{i=1}^{N}(||z_{where_{i, t}}\!-\!\widehat{z_{where_{i, t}}}||^2\!\cdot \mathds{1}_{z_{pres_{i, t}} > 0.5} \\
& + ||z_{where_{i, t}}\!-\! \widehat{z_{where_{i, t}}}||^2\!\cdot \mathds{1}_{\widehat{z_{pres_{i, t}}} > 0.5}).\notag
\end{align} 
We combine these losses to a motion supervision loss:
\begin{equation}
\mathcal{L}^\text{M} := \lambda_\alpha \cdot \mathcal{L}_\alpha + \lambda_{pres} \! \cdot \! \mathcal{L}_{pres} + \lambda_{where} \cdot \mathcal{L}_{where},
\end{equation}
where $\lambda_\alpha$, $\lambda_{pres}$ and $\lambda_{where}$ weigh the loss components. 

Further, to bootstrap the performances of the decoder, we introduce a guidance mechanism on the variables $z_{where}$, $z_{pres}$ and $\alpha$. For each of these variables, $v$, we provide a convex combination, $v = \lambda_\text{guid} \cdot \hat{v} + (1 - \lambda_\text{guid}) \cdot v$ of the model's prediction and the ``ground truth'' information from motion estimation to the decoder. This mechanism acts as a kind of mixing of motion guidance and allows us to improve the image reconstruction in early steps (via altering $\alpha$ and $\texttt{fg}$ in Eq.~\ref{eq:reconstruction}), and with $\lambda_\text{guid}$ to progressively decay its influence in the latter course of training (\cf App.~\ref{appendix:hyperparameters}).

For detected objects in consecutive frames, we apply a contrastive loss, $\mathcal{L}^\text{OC}$, on the internal representations (\ie $\texttt{enc}$) of the model. This loss makes the representations of the same object depicted over several frames similar, and the representations of different objects heterogeneous. We estimate whether two objects in consecutive frames represent the same entity based on their \texttt{loc} and \texttt{enc} variables.

For $\mathcal{L}^\text{OC}$ we denote $\Omega_{t} = \{{o}_{t}^j\}_{j=1}^{c_t}$ as the set of the $c_t$ detected object representations (\ie for which $z_{pres}>0.5$) in the frame $\textbf{x}_t$. 
Let $\texttt{enc}^*({o}_{t}^j)$ denote a function providing the most similar object to ${o}_{t}^j$ of $\Omega_{t\!+\!1}$, based on the cosine similarity ($S_C$) of the encodings. Correspondingly, $\texttt{loc}^*({o}_{t}^j)$ is a function that provides the nearest object to ${o}_{t}^j$, of $\Omega_{t\!+\!1}$, based on the locations (via Euclidean distance).
If these functions return the same object, we can safely assume that it corresponds to the same object (as $o_{t}^j$) in the next frame. We thus introduce the contrastive object continuity loss as:

\begin{equation}
\mathcal{L}^\text{OC}(\textbf{x}_{t}) = \sum_{o^{i}_{t} \in \Omega_{t}} \sum_{o^{j}_{t\!+\!1} \in \Omega_{t\!+\!1}} \lambda_{\text{differ}} \cdot S_C(o^{i}_{t}, o^{j}_{t\!+\!1}),
\end{equation}
where $\lambda_{\text{differ}}$ is defined as:
\begin{equation}
    \lambda_{\text{differ}} =
    \begin{cases}
        -5 & \text{if } o_{t+1}^j = \texttt{enc}^*({o}_{t}^i) = \texttt{loc}^*({o}_{t}^i) \\
        1 & \text{else} \\
    \end{cases}
\end{equation}
This computation can notably be optimized for grid based models by focusing $\Omega_{t}$ only on a square grid around the object, \eg a $3 \times 3$ grid, thus reducing the computational cost (\cf App.~\ref{app:loc_optim}). 

\subsection{MOC Details for Slot-Attention Models}
\label{appendix:samoc}
An introduction of the notations for Slot Attention used hereafter are provided in App.~\ref{appendix:slot}.

In our experiments, the input dimension is $128 \times 128 \times 3$, and the internal binary masks dimension is $32 \times 32$. The optical flow (OF) masks $\{\alpha_i\}_{i=1}^{\hat{c}}$ are created from the original $128 \times 128$ masks, using the \texttt{block\_reduce} function of scikit-learn image\footnote{\url{https://scikit-image.org/docs/stable/api/skimage.measure.html\#skimage.measure.block\_reduce}} with max as the block-applied function.

We adapt the motion loss of MOC for Slot Attention as was done in \cite{Bao2022discovering}. It consists in searching for the permutation of L elements with the lowest cost, corresponding to a bipartite matching between the $L$ internal masks of the model and the ones obtained from OF:
\begin{equation}
\hat{\sigma}=\underset{\sigma}{\arg \min } \sum_{i=1}^L \mathcal{L}_{s e g}(\hat{\alpha}_i^t, \alpha_{\sigma(i)}^t)    
\end{equation}
and use it to compute the motion loss of slot attention
\begin{equation}
\mathcal{L}^{\text{M}}(\textbf{x}_{t})=\sum_{i=1}^{L} \mathds{1}_{\{\hat{\alpha}_i^t \neq \emptyset\}} \mathcal{L}_{s e g}(\hat{\alpha}_i^t, \alpha_{\hat{\sigma}(i)}^t).
\end{equation}

Finally, $L_{seg}$ is defined as the binary cross entropy:
\begin{equation}
\mathcal{L}_{s e g}(\hat{\alpha}, \alpha) = \sum_{j=1}^{N} -\hat{\alpha}_j \text{log}(\alpha_j) - (1 - \hat{\alpha}_j) \text{log}(1-\alpha_j).
\end{equation}

Note that because our model can have more slots than there are objects in the image (\ie $L > c^t$), we can define empty masks for $\{\alpha^t_i\}_{i=c^t+1}^{L}$.

For the OC loss we denote $\Omega_{t} = \{{o}_{t}^j\}_{j=1}^{c_t}$ as the set of the objects in the frame $\textbf{x}_t$ with ${o}_{t}^j = (\textbf{z}_{t}^j, \alpha_{t}^j)$. 
For $\texttt{enc}^*$ we next compute the cosine similarity between slot encodings and choose the most similar encoding in the consecutive frames from $\Omega_{t+1}$ for $o_{t}^j$. For $\texttt{loc}^*$ we use $L_{s e g}$ between each  $\alpha_{t}^j$ of $\Omega_{t}$ and each $\alpha_{t+1}$ of the consecutive frame, $\Omega_{t+1}$), providing us with a score of similar objects based on the location representation, i.e. slot attention masks. From this score we identify the best matching for each $\alpha_{t}^j$. If these two functions return the same object id, we can safely assume that the identified object in frame $t+1$ corresponds to the same object (as $o_{t}^j$). 

$\mathcal{L}^\text{OC}$ is finally computed as: 
\begin{equation}
\mathcal{L}^\text{OC}(\textbf{x}_{t}) = \sum_{o^{i}_{t} \in \Omega_{t}} \sum_{o^{j}_{t\!+\!1} \in \Omega_{t\!+\!1}} \lambda_{\text{differ}} \cdot S_C(o^{i}_{t}, o^{j}_{t\!+\!1}),
\end{equation}
where $\lambda_{\text{differ}}$ is defined as:
\begin{equation}
    \lambda_{\text{differ}} =
    \begin{cases}
        -\beta & \text{if } o_{t+1}^j = \texttt{enc}^*({o}_{t}^i) = \texttt{loc}^*({o}_{t}^i) \\
        1 & \text{else} \\
    \end{cases}
\end{equation}

In all of our experiments, for both models, we use $\beta = 5$.

\subsection{Details on Scheduling}
\label{appendix:scheduling}
We propose a dynamic scheduling for SPACE+MOC based on the model's prediction performances instead of a constant decay. Specifically, we compare the amount of detected objects by the model ($c = \sum \round{z_{pres}}$) and optical flow ($\hat{c} = \sum \round{\widehat{z_{pres}}}$). We also compare the detected bounding boxes (deduced from $\texttt{loc}$) between the model's predictions and the motion estimates and summarize these into a single heuristic value. 
We denote $\mathds{BB}_{t}$ as the set of bounding boxes, constructed from $\texttt{loc}$ (detailed for SPACE in Appendix \ref{algo:zwheretobbox}) of all the objects detected in a single frame (\ie for which $z_{pres}>0.5$) and compute a bounding box matching score, defined as:
\begin{equation}
\delta_\text{BBMS} := \!\sum_{bb \in \mathds{BB}_{t}} \min_{\hat{bb} \in \mathds{\hat{BB}}_{t}}  MSE(bb, \hat{bb})
\end{equation}
where $\hat{\mathds{BB}}_{t}$ represents the set of detected objects via the motion mechanism of section \ref{sec:motion}.  $\delta_\text{align}$ is finally computed using positive parts and tolerances on these metrics:
\begin{equation}
\label{eq:delta_align}
\delta_\text{align} := (\delta_\text{BBMS} - \beta_\text{mismatch})^+ + ( c - \hat{c} \cdot \beta_\text{underestimation})^+
\end{equation}
As the motion mechanism systematically underestimates the number of objects (motion cannot detect stationary objects), we allow the model to detect up to $\beta_\text{underestimation}$ times the objects without intervention. Secondly, we introduce a tolerance, $\beta_\text{mismatch} \in \mathbb{R}^+$, for the bounding box matching score for object location variances.
Both of these represent hyperparameters of the MOC framework (details given in the Appendix~\ref{appendix:hyperparameters} for chosen values).
Lastly, we express $\lambda_\text{align} := 2^{-\delta_\text{align}}$. Note that we constrain $\delta_\text{align} \in \mathbb{R}^+$ to enforce $\lambda_\text{align} \in [0, 1]$.
$\lambda_\text{align}$ is computed on a batch of frames once at the start of every epoch.

For SLAT+MOC we use a simpler scheduling approach consisting of a linear increase of $\lambda_\text{align}$ between $0$ and $1$ up to $\frac{n_{epochs}}{2}$. 

\subsection{Computing distances between detected objects}
\label{algo:zwheretobbox}
In order to compute different metrics, we need to compute the bounding boxes that correspond to different $z_{where}$  variables, as the $z_{where}$ variables encode the center position and the size of the objects. The following function is used to convert the variables of SPACE in bounding boxes:

\lstset{style=pythonstyle}

\begin{lstlisting}[language=Python, caption=Code to convert $z_{where}$ to bounding box, linewidth=14cm]
def zwhere_to_box(z_where):
    width, height, center_x, center_y = z_where
    center_x = (center_x + 1.0) / 2.0
    center_y = (center_y + 1.0) / 2.0
    x_min = center_x - width / 2
    x_max = center_x + width / 2
    y_min = center_y - height / 2
    y_max = center_y + height / 2
    
    pos = [y_min, y_max, x_min, x_max]
    return pos
\end{lstlisting}

To compute $\delta_{align}$ in eq.~\ref{eq:delta_align}, we apply the Mean Square Error functions to bounding boxes converted with the previous code from the $z_{where}$ latent variables. 

\subsection{Fast \texorpdfstring{$\mathcal{L}^\text{OC}$}{TEXT} computation for SPACE+MOC}
\label{app:loc_optim}
For grid based object detection models such as SPACE, the computation of $\mathcal{L}^\text{OC}$ in Eq.~\ref{eq:loc_def} can be optimized via looking only at the objects on the $3\times3$ grid around the object. For an object ${o}_{t}^j$, we then denote $\Omega^{3 \times 3}_{t\!+\!1,j}$ as the set of all detected objects in a $3\!\times\!3$ square kernel around this cell at time $t\!+\!1$.
Let $\texttt{enc}^*({o}_{t}^j)$ denote the function providing the most similar object to ${o}_{t}^j$ among the objects of this surrounding grid ($\Omega^{3 \times 3}_{t\!+\!1,j}$), based on the cosine similarity ($S_C$) of the encodings, and now correspondingly, $\texttt{loc}^*({o}_{t}^j)$ is the function that provides the nearest object to ${o}_{t}^j$, of $\Omega^{3 \times 3}_{t\!+\!1, j}$, based on the locations (via Euclidean distance).
We thus can compute $\mathcal{L}^\text{OC}$ with the updated following equation:
\begin{equation}\label{eq:oc_single_3x3}
\mathcal{L}^\text{OC}(\textbf{x}_{t}) = \sum_{o^{j}_{t} \in \Omega_{t}} \sum_{o^{j}_{t\!+\!1} \in \Omega^{3 \times 3}_{t\!+\!1,j}} \lambda_{\text{differ}} \cdot S_C(o^{j}_{t}, o^{j}_{t\!+\!1}).
\end{equation}
Again, denoting ($c_t \leq H \times W)$ as the number of detected objects in a frame, we go from a computational complexity of $\mathcal{O}(c_t \times c_{t+1}) \approx \mathcal{O}(c_t^2)$ to $\mathcal{O}(9 \times c_t)$.

\newpage
\section{Additional Visual results}
\label{appendix:visual_results}
This section provides additional results that complete the ones of the main paper. In the following, we provide detailed evaluation for each tested environment, and each tested model:
the evolution of the validation F-scores during training, Precision/Recall curves, the evolution of Few-shot accuracies with growing number of samples, Final Few-shot accuracies on relevant objects and Final F-scores and Adjusted Mutual Information scores on all objects. For this section, we denote SPOC as SPACE + MOC, the SPACE baseline detection model trained with Motion Supervison and Object Continuity.

\subsection{Evolution of F-scores during training}
\vspace{-0.5cm}
\begin{figure*}[ht]
    \centerline{\includegraphics[width=1.\textwidth]{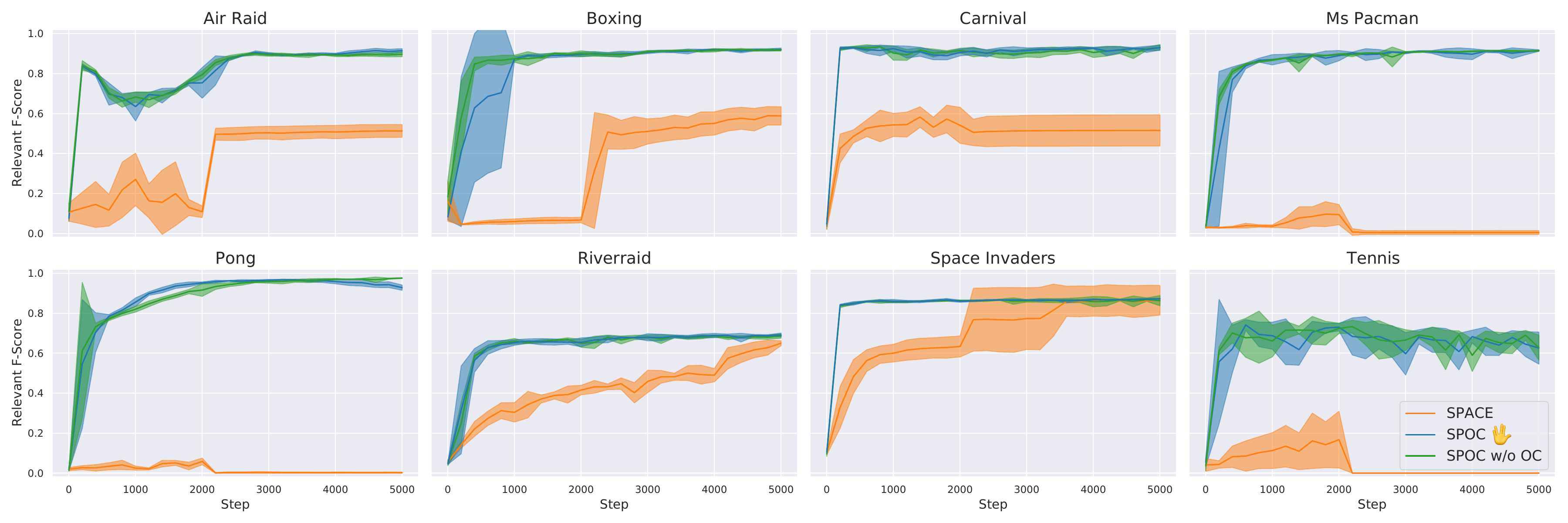}}
  \caption{Object priors via motion enables faster convergence. Each subfigure presents the validation F-score over training on the validation set for each Atari environment. Our simple optical flow approach is not optimized for the moving backgrounds of Riverraid and Tennis, performances could be further improved with a better one. Figure best viewed in color.}
  \label{fig:f_sco_overtime} 
\end{figure*}

\subsection{Precision Recall Curves}
\vspace{-0.5cm}
\begin{figure*}[ht]
    \centerline{\includegraphics[width=1.\textwidth]{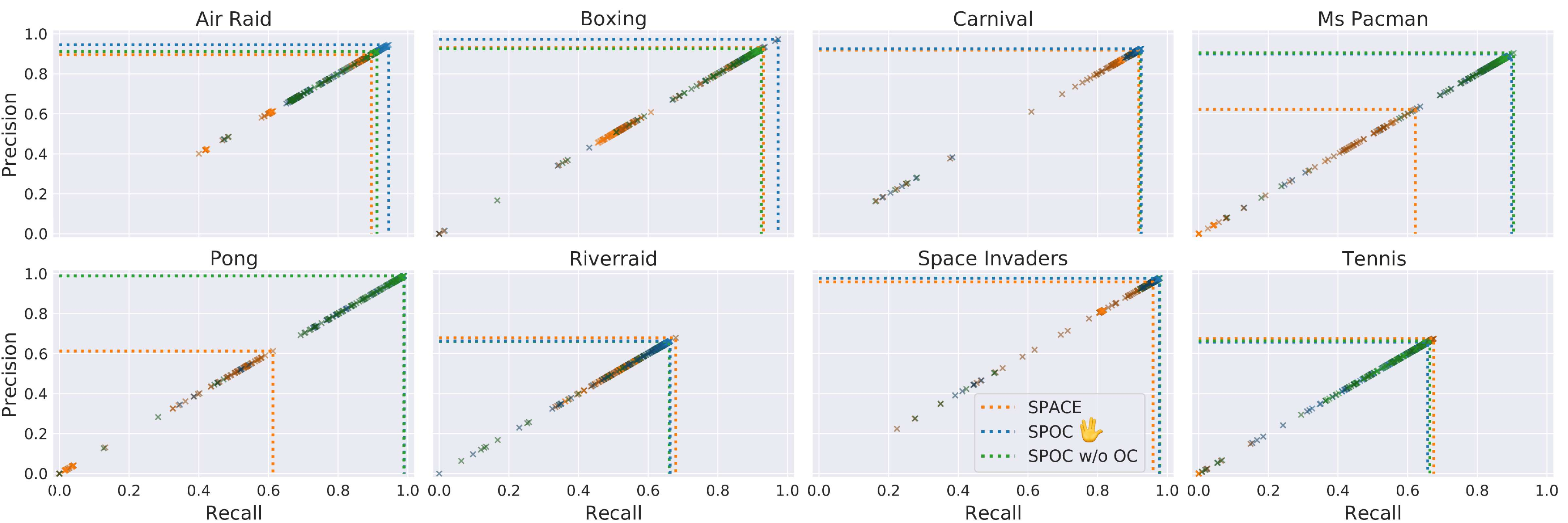}}
  \caption{Time-based supervision allows achieving both higher Precision and Recall at once, and reaches a high absolute Precision in general. Each cross marker corresponds to an evaluation result during training from any seed. Lighter colors symbolize a result towards the end of training. Furthermore, we linearly interpolate the Pareto Boundaries of all measurements for all models (dotted lines).}
  \label{fig:pr_curves}
\end{figure*}

\newpage
\subsection{Few-shots accuracies with growing number of samples.}
\label{appendix:ablation_fsa}
\begin{figure*}[ht]
    \centerline{\includegraphics[width=1.\textwidth]{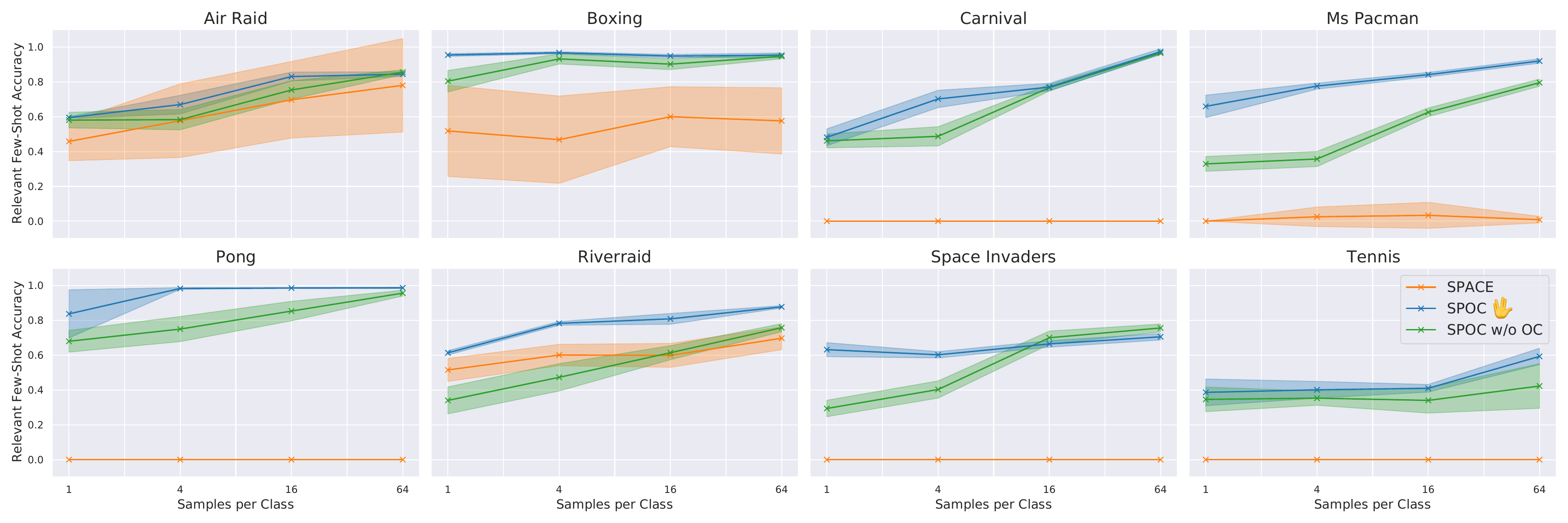}}
  \caption{MOC allows for fewer shot learning. The evolution of few-shot classification learners with varying amount of samples (per class) provided to them, based on the encodings of the objects detected by SPACE, SPACE + MOC w/o OC, and the full SPACE + MOC models. Figure best viewed in color.}
  \label{fig:fsa_overtime}
\end{figure*}

\subsection{Final Relevant Few shot accuracy with 4 and 16 samples}
We here present the relevant (\ie only objects that are important to play the game, i.e. no score, HUD, etc.) few shot accuracy ($1$, $4$, $16$ and $64$ samples of each object class) on each tested game.
\begin{figure}[ht]
    \centering
    \includegraphics[width=0.97\textwidth]{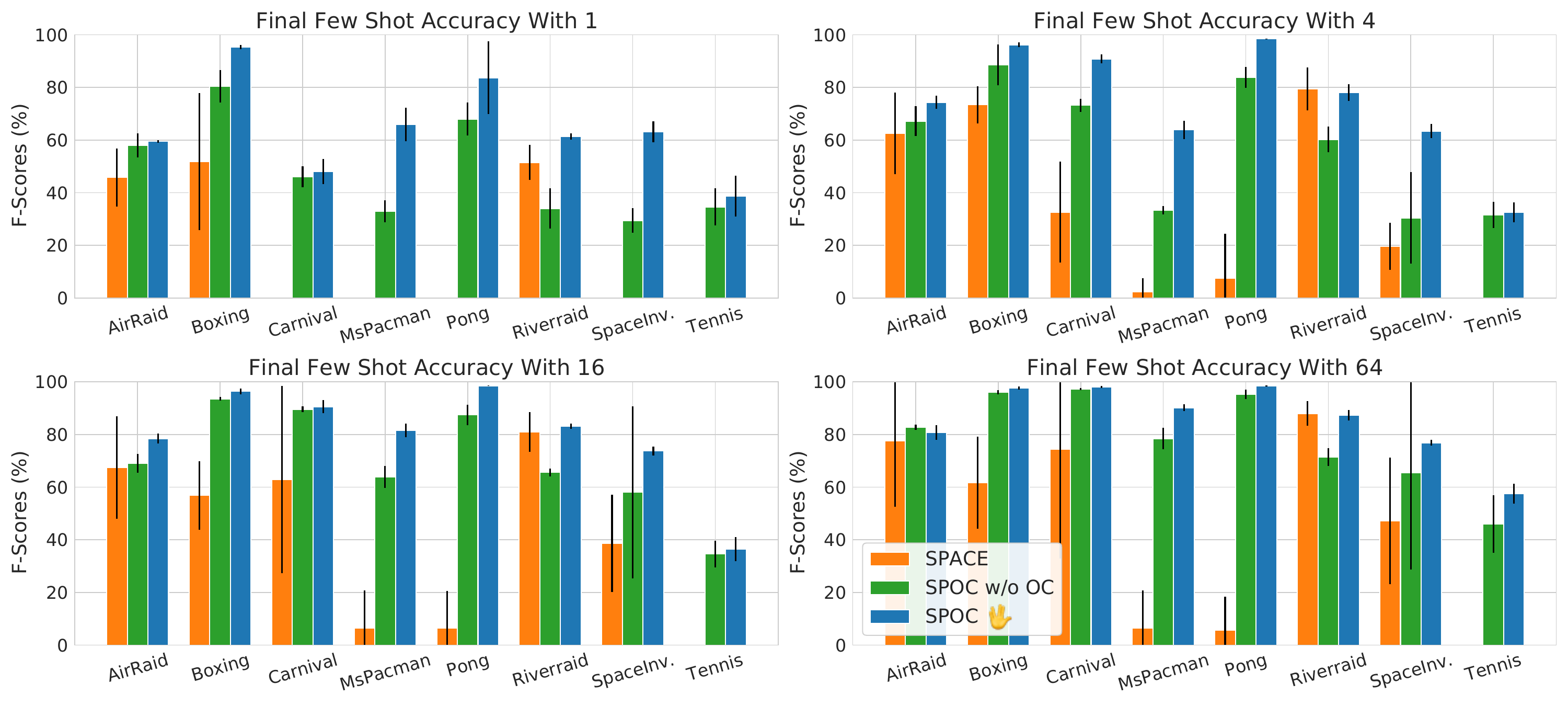}
    \caption{Motion supervision together with Object Continuity of our MOC framework leads to more optimal object encodings. The relevant few shot accuracy, with $1$ (top left), $4$ (top right), $16$ (bottom left) and $64$ (bottom right) samples of each class, of our different model setups. The model with Object Consistency always obtains the best average performances. Figure best viewed in color.}
    \label{fig:few_shot_acc_4_16} 
\end{figure}

\newpage
\subsection{Final F-scores and Adjusted Mutual Information on all objects.}
\label{appendix:ablation}
In our paper, we focus our evaluations on \textit{relevant} object, \ie objects that are necessary or useful for the agents' policy. Hereafter, we also provide F-scores and Adjusted Mutual Information scores evaluations on all objects, including ones of the HUD, such as scores and lives.

\begin{figure}[ht]
    \centering
    \includegraphics[width=0.97\textwidth]{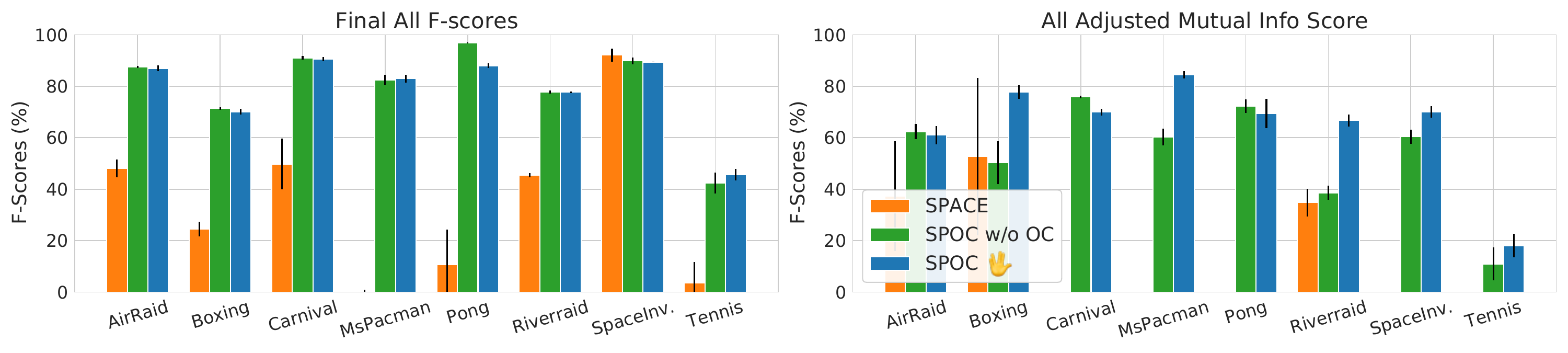}
    \caption{Performances improvement on object detection and encoding are also valid when evaluating on every object of the environments. F-scores (left) and Adjusted Mutual Information scores (right) of the SPACE baseline models, SPACE + MOC w/o OC, and full SPACE + MOC (SPOC). Figure best viewed in color.}
    \label{fig:fsa_ami_all} 
\end{figure}

\subsection{Qualitative results}
\label{sec:qualitative}
Here we provide qualitative results of the SPACE model, SPACE with time supervision (SPACE + MOC w/o OC) as well as with both motion supervision and object continuity (SPACE + MOC, abbr. SPOC). 
\begin{figure}[ht]
    \centering
    \includegraphics[width=1.\textwidth]{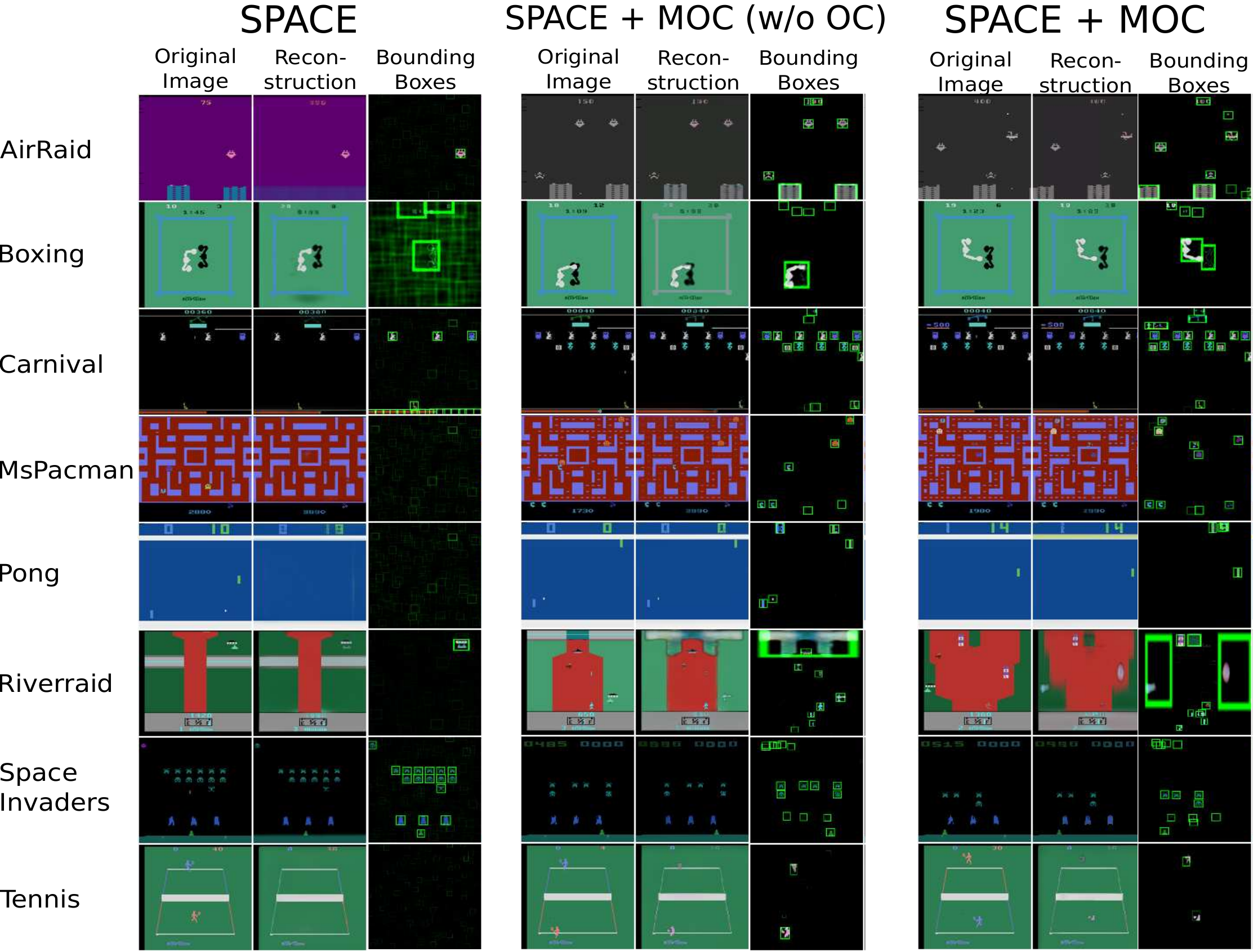}
    \caption{Performances improvement on object detection and encoding are also valid when evaluating on every object of the environments. F-scores (left) and Adjusted Mutual Information scores (right) of the SPACE baseline models, SPACE + MOC w/o OC, and full SPACE + MOC. Figure best viewed in color.}
    \label{fig:models_qual_comp} 
\end{figure}

\subsection{PCA of embedding latent spaces of the object representations across models}
\label{figs:PCA_enc}
To reduce the size of the PDF and its loading time, the following section is removed; only available in previous Arxiv versions of this manuscript.

\newpage
\section{Hyperparameters}
\label{appendix:hyperparameters}
\subsection{SPACE Hyperparameters}
This section provides details about how we computed in practice our object continuity contrastive loss, as well as hyperparameters of our loss framework and of our framework and on the SPACE original architecture (if modification were needed). 
Table~\ref{table:basehyper} presents the default hyperparameters used for SPACE + MOC and Table \ref{table:origin} displays the differences between games. For switching between SPACE + MOC w/o OC and SPACE + MOC $\lambda_\text{OC}$ (the weight of object continuity) is set to $0$ or $10$ respectively. Hyperparameters were visually evaluated on a separate evaluation set. 
For the original hyperparameters and the architecture of SPACE, please also refer to \cite{SPACE2020}. 

\begin{table}[!ht]
\setlength{\tabcolsep}{5pt}
\centering
\begin{tabular}{@{}ll@{}}
\toprule
\textbf{Parameter} & \textbf{Value} \\ 
\midrule
Foreground / Background lr & $3 \cdot 10^{\text{-}5}$ / $10^{\text{-}3}$ \\
batch size & $16$ \\
gradient steps & $5000$ \\
\midrule
$z^{pres}$ prior probability & $0 \rightarrow 5000 : 0.1 \rightarrow 10^{\text{-}10}$ \\
$z^{scale}$ prior mean & $0 \rightarrow 5000 : \text{-}2 \rightarrow \text{-}2.5$ \\
$z^{scale}$ prior std & $0.1 \cdot I$ \\
$z^{shift}$ / $z^{what}$ / $z^{depth}$ priors & $\mathcal{N}(0, I)$ / $\mathcal{N}(0, I)$ / $\mathcal{N}(0, I)$ \\
$\tau$ (gumbel-softmax-temperature) & $2.5$ \\
Foreground / Background stds & $0.2$ / $0.1$\\
Background Components & $3$ \\
Grid Size & $16$ \\
fixed $\alpha$ \& boundary loss & removed \\
\midrule
Motion Kind & Mode \\
$\eta$ & $0.5$ \\
$\lambda_\alpha / \lambda_{pres} / \lambda_{where}$ & $100$ / $1000$ / $10000$ \\
$\lambda_\text{guid}$ & $0 \rightarrow 3000 : 1.0 \rightarrow 0.0$ \\
$\beta_\text{mismatch}, \beta_\text{underestimation}$ & $0.1$, $1.25$ \\
\bottomrule
\end{tabular}
\vspace{0.2cm}
\caption{Default hyperparameters used for SPACE models and our Motion and Object Continuity training scheme.}
\label{table:basehyper}
\hfill
\end{table}

\begin{table}
\centering
\begin{tabular}{@{}ll@{}}
\toprule
\textbf{Parameter} & \textbf{Value} \\ 
\midrule
Boxing & \\
\hspace{5mm}$z^{scale}$ prior & -1 \\

\midrule
MsPacman & \\
\hspace{5mm}$z^{scale}$ prior mean & $1K \rightarrow 5K : 0.1 \rightarrow 10^{\text{-}10}$ \\
\hspace{5mm}$\beta_\text{underestimation}$ & $1.5$ \\

\midrule
Pong & \\
\hspace{5mm}$\beta_\text{underestimation}$ & $1.5$ \\

\midrule
Riverraid & \\
\hspace{5mm}BG Components & $4$ \\
\hspace{5mm}$\beta_\text{underestimation}$ & $1.5$ \\
\hspace{5mm}$\lambda_\text{mismatch}$ & $0.1$ \\

\midrule
Tennis & \\
\hspace{5mm}$z^{scale}$ prior mean & $0 \rightarrow 5K : \text{-}2.5 \rightarrow \text{-}3.0$ \\
\hspace{5mm}$\beta_\text{underestimation}$ & $0.75$ \\
\hspace{5mm}$\lambda_\text{mismatch}$ & $0.01$ \\

\bottomrule
\end{tabular}
\vspace{0.2cm}
\caption{Deviations from the default Hyperparameters of SPACE + MOC for specific games.}
\end{table}

\begin{table}
\setlength{\tabcolsep}{6pt}
\centering
\centerline{
\begin{tabular}{@{}ccccc@{}}
\toprule
\textbf{Metric} & \textbf{Game} &  \textbf{Origin}   & \textbf{SPACE}   & \textbf{SP.+MOC}  \\ \midrule
\multirow{2}{*}[0.2em]{\textbf{F-Score}} & Riverraid                      & 27.1\tiny$\pm$26.0         & 65.0\tiny$\pm$1.2        & \textbf{69.1\tiny$\pm$1.0}               \\
& SpaceInv.                 & 76.5\tiny$\pm$10.2         & \textbf{86.5\tiny$\pm$7.4}        & \textbf{87.2\tiny$\pm$0.3} \\
\midrule
\multirow{2}{*}[0.2em]{\textbf{Mutual Info.}} & Riverraid                      & 37.0\tiny$\pm$33.9       & 41.1\tiny$\pm$3.5      & \textbf{60.8\tiny$\pm$1.0}               \\
& SpaceInv.                 & 45.8\tiny$\pm$26.1       & n/a      &  \textbf{76.7\tiny$\pm$2.6} \\
\midrule
\multirow{2}{*}[0.2em]{\textbf{FSA}} & Riverraid                      & 45.0\tiny$\pm$41.2       & 60.1\tiny$\pm$6.1      & \textbf{78.3\tiny$\pm$1.1}               \\
& SpaceInv.                 & 25.8\tiny$\pm$15.2       & n/a      & \textbf{60.2\tiny$\pm$1.8} \\
\bottomrule
\end{tabular}
}
\vspace{0.2cm}
\caption{The higher learning rate used during the evaluations for the baseline does not harm performances. We compare the metrics between SPACE with the original learning rate (and 10k steps (\textbf{Origin}), Baseline SPACE with a higher learning rate and 5k steps (\textbf{SPACE}), and \textbf{SPACE + MOC} with the higher learning rate and 5k steps. The result indicates that the higher learning rate did not reduce performances, neither in terms of $F_1$, Adjusted Mutual Information, or with Linear Classification Accuracy (trained with 4 samples per class). }
\label{table:origin}
\end{table}

\subsection{Original SPACE Hyperparameters}
\label{appendix:origin}

SPACE authors \cite{SPACE2020} reported their hyperparameter settings for the games Riverraid and Space Invaders. For fair evaluation, we reused those configurations, but increased the foreground learning rate (from $10^{-5}$ to $3 \cdot 10^{-5}$) as we noticed the lower learning rate leads to a long training time. In Table~\ref{table:origin} we present the effect of varying learning rates evaluated via our three main metrics: adjusted mutual information, few-shot classification accuracy and f-score. As can be seen, our choice of learning rate is in fact favorable, even though less training steps are required. 

\subsection{Slot Attention Details and Hyperparameters}
\label{appendix:slat_hp}
For Slot Attention, we use a Residual Network \textit{ResNet18}~\cite{HeZRS16}. The differences of our model and the original Slot Attention ones are the learnable slot initialization and the first frame pad. All hyperparameters are the same across the different games, for both models with and without MOC supervision. The hidden dimension is $64$. We use the same dimension for the flow masks for the supervision. We use a batch size of $18$, for $50$ epochs only, $5000$ max steps. $\beta$ was set to -5.
We used $\lambda_{OC}=1.0$ for the games Pong, Carnival and Spaceinvaders and $\lambda_{OC}=0.1$ for the remaining games of Atari-OCTA. $\lambda_{mask}=1$ for all games.

The number of slots for each game are however different, as they all contain different maximum amount of objects. We provide the used number hereafter:

\begin{table}[!ht]
\centering
\setlength{\tabcolsep}{4pt}
\begin{tabular}{lcccccccc}
Game   & Airraid & Boxing & Carnival & MsPac. & Pong & Riverraid & Sp.Inv. & Tennis \\ \hline
SLOTS & 20       & 10     & 50       & 20       & 10   & 20        & 50         & 10    
\end{tabular}
\end{table}

For our Slot Attention evaluations we provide statistics over 3 random seed initializations.

\newpage
\section{Evaluation Metrics}
\label{appendix:evaluation_metrics}
In this section, we provide details about the different evaluation metrics used across the paper.

\subsection{Precision and Recall}
A reliable object detection algorithm for a downstream task should both detect every depicted object and avoid returning non-objects (\ie when background elements are reconstructed as objects). The former point is captured by the precision, defined as $P=\frac{\text{\#detected true objects}}{\text{\#detected objects}}$. The share of correctly determined objects, among all those found. The recall, defined as $R=\frac{\text{\#detected true objects}}{\text{\#true objects}}$, highlights how many of the desired objects were actually found.

\begin{figure}[ht]
  \centering
  \begin{subfigure}{.4\textwidth}
  \captionsetup{justification=centering}
    \includegraphics[width=\linewidth]{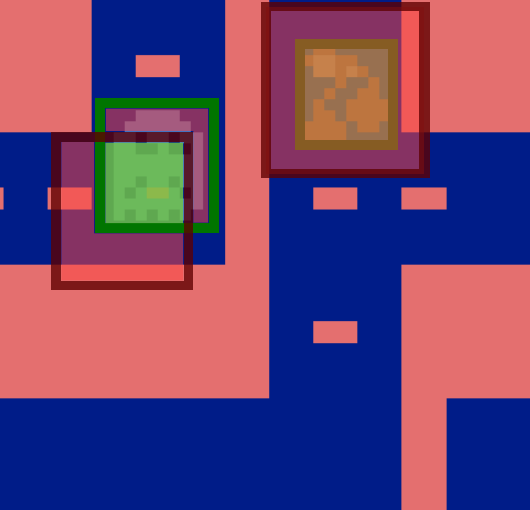}
    \caption{Intersection over Union (IoU) computes the ratio between the green and the red areas.}
    \label{fig:iou}
  \end{subfigure} 
  \begin{subfigure}{.538\textwidth}
  \captionsetup{justification=centering}
    \includegraphics[width=\linewidth]{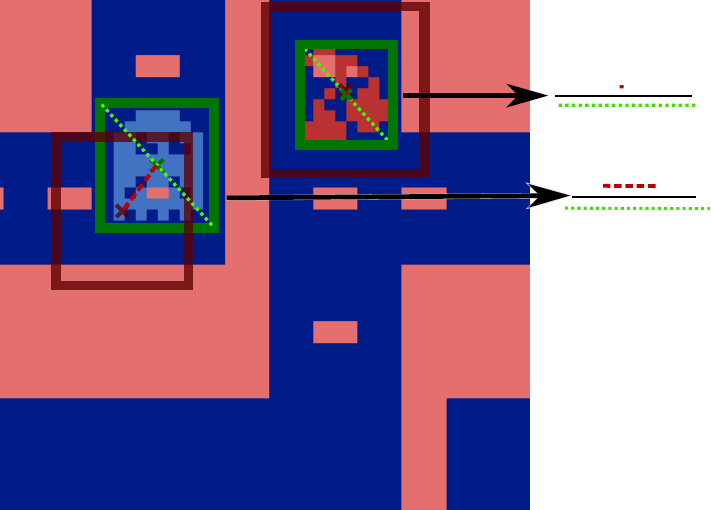}
    \caption{Center Divergence is computing using the ratio between the boxes' center points and the object size.}
    \label{fig:center_div}
  \end{subfigure}
  
  \caption{\small{Center Divergence accurately models a condition for successful detection from the RL viewpoint.
  This can be seen in the direct comparison to the other introduced metrics for bounding box comparison. IoU considers both bounding boxes and thus might disregard the fruit in this example, while IoU* considers only the covered share of the ground truth and does not react to the misaligned bounding box for the ghost. Finally, Center Divergence captures both notions, by evaluating the center distances in relation to the size of the objects.}}
  \label{evaluation::prec_recall} 
\end{figure}

To assess that a detected object correspond to a true object, a classic metric is the Intersection over Union (IoU), defined as the amount of intersecting area, over the total area covered by two areas (\cf Figure~\ref{fig:iou}). One considers an object to be correctly detected if the bounding box of the object found by the model has a sufficiently high IoU with the ground truth bounding box. 
In detail, for each predicted bounding box, we search for the matching ground truth bounding box with the largest overlap, threshold by some IoU value, remove twice matched predicted bounding boxes, and count the number of hits for precision and recall computation. With a growing IoU threshold, the number of detected objects decreases, and the precision thus augments while the recall decreases.

Furthermore, an object detection algorithm can be considered superior if the given bounding boxes are matching well to ground-truth bounding boxes, which leads to the classic metric of Average Precision (AP) commonly used in the works we build on \cite{locatello2020slotattention, SPACE2020}. The general definition for the Average Precision (AP) is finding the area under the precision-recall curves (given for our methods and datasets in Figure~\ref{fig:pr_curves}). Commonly, an 11-point interpolated average recall levels ($r \in \{0.1 \cdot n \; | \; n \in [0, 10] \cap \mathbb{N}\}$) is used, for which the corresponding precision are retrieved and averaged. 
In addition to the average of the 11-points, we average those values over similarly distributed IOU thresholds to compute a singular metric, attending to the quality of the determined bounding boxes. We report AP of our $3$ evaluated models in the Appendix~\ref{appendix:results_ap}. 

When considering practicalities, we believe this assessment of bounding boxes to be slightly unintuitive and misleading.
Firstly, confidence-based ordering is barely sensible in practice if SPACE usually assigns high confidence to all its objects as it is not trained towards being uncertain (like in Open Set Recognition \eg \cite{scheirer2012toward}). To circumvent this practical issue, we simplified our assessment to Precision and Recall over \textit{all} found objects (\ie for which $z_{pres} > 0.5$).
Secondly, and more importantly, SPACE bases its bounding boxes' size predictions on a Gaussian prior. Thus, it predicts bounding boxes with similar sizes for every object. It thus might need to choose significantly larger bounding boxes for smaller objects in games with objects of varying sizes, such as Pong and Tennis. However, predicting bigger bounding boxes shouldn't be penalized in our case, also cause object sizes are fixed for each object classes, and can thus be implicitly learned by the policy. Furthermore, one could use the prior given by Motion Supervision to parameters of a Mixed Gaussian distribution.

In practice, we observed cases, where even with an IOU threshold of 0.2 the object was not considered as recognized, although we perceive it as detected, as it contributes solely the object to the reconstruction. As an intuitive example, consider a bounding box that is trice as large as the ground truth, which leads to the low IOU of $\frac{1}{9}$. This problem is intensified by the precise automated object labeling, which will not provide any border around the object, that might be required to capture pixels influenced through interpolation when resizing the image.

Accordingly, we introduce Center Divergence (\cf Figure~\ref{fig:center_div}) to replace IoU. Center Divergence describes the distance between bounding boxes' center points, normalized by the actual object size (length of bounding box diagonal). In practice, we used a threshold of $0.5$ on the center divergence to decide whether a detected box corresponds to a ground truth object. Center divergence is used for the Precision and Recall as it matches the expectation of humans and symbolic object-centric RL algorithms, that an object is given by its position and type (or class) and not necessarily the full bounding box. 

Those metrics, though meaningful on their own, are commonly combined into the single value, the F-Score, defined as the harmonic mean of Precision and Recall. In short, our computed F-scores utilizes Center Divergence then presents a suitable metric for judgment of object detection and avoids the unintuitive pitfalls of alternative metrics for the symbolic RL context.

\subsection{Adjusted Mutual Information}

Among other things, with our framework we aim to improve the encoding representation for incorporation in downstream tasks such as reinforcement learning. One way to evaluate the quality of the encoding space is by investigating if the encoding space contains semantic meaningful subspaces, e.g. that correspond to underlying object classes. 

One naive approach is to perform a simple cluster evaluation of the latent object representations and measure how well the clustering corresponds to a ground truth clustering. In order to provide a ground truth clustering to compare to we use IOU* as described above to assign object labels to each encoding (with an additional "no-label"-label, for predicted bounding boxes, that do not correspond to any object). In other words we map a nearest ground truth object to each predicted object encoding giving us an estimate of the encodings object classes. 

Next, we compute a simple clustering of the latent object encodings using K-Means (where k=\#Classes) and evaluate the quality of the clustering using the (Adjusted) Mutual Information, which in its unadjusted form for two clusterings $C_1: X \longrightarrow Y_1$ and, $C_2: X \longrightarrow Y_2$ is defined as:

{
\fontdimen16\textfont2=3pt
\fontdimen17\textfont2=3pt
\begin{align*}
    \operatorname{MI}(C_1, C_2) &= D_{\mathrm{KL}}( P_{(C_1,C_2)} \| P_{C_1} \cdot P_{C_2} ) \\
    &= \sum_{y_1 \in Y_1} \sum_{y_2 \in Y_2}
    { p_{(C_1,C_2)}(y_1, y_2) \log\left(\frac{p_{(C_1,C_2)}(y_1, y_2)}{p_{C_1}(y_1)\,p_{C_2}(y_2)}\right) }
\end{align*}

where $D_{\mathrm{KL}}$ describes the KL-Divergence and the probabilities follow empirically from the share of assignment to each class (e.g. $p_{C}(a) = \frac{|\{x \in X \; | \; C(x) = a\}|}{|X|}$ ). 
}



We choose Mutual Information (MI) following \cite{romano2016ami}, which supports its usage with the unbalanced classes present in Atari games.
It is adjusted by the expected MI of a random clustering \cite{vinh2010adjusted} such that the space between 0 and 1 spans the range from a random cluster assignment to the optimal clustering while maintaining its expressive intuition.

In summary, we present AMI as one measure for the quality of an encoding space, by exhibiting the similarity of a simplistic clustering to the ground truth labeling. 

\subsection{Classifiers}
\label{eval::classifier::section}

In addition to AMI, the performance of a few-shot classifier can provide information on the quality of a model's encoding space in the sense that high classification accuracy (particularly of a linear classifier) indicates that the encoding space is well separable. Thus, this can act as a proxy estimation for the performance of other downstream tasks. If a simple classifier can distinguish the encoding space it seems likely that also an RL agent can learn form the encodings. Whereas if the classifier already fails it seems likely also an RL agent will fail. 



In this work we revert to simple ridge regression \cite{hoerl1970ridge1, hoerl1970ridge2} classifiers. Specifically, we train a ridge regression classifier on few encodings per class ($n \in \{1, 4, 16, 64\}$ to be precise) to measure the encoding quality with linear probing in a few shot setting. We test the classifier on a separate test set. The hyperparameters of the ridge regression model were set as the default values of the sklearn package\footnote{\url{https://scikit-learn.org/stable/modules/generated/sklearn.linear_model.Ridge.html}}. The number of classes was dependent on the Atari-OCTA game and the object encodings were balanced across classes. 

\subsection{Additional Technical Evaluation Details}

Finally, we report the details and settings used for evaluating our methods. 

The model is implemented in PyTorch 1.9.0 and evaluation is supported by sklearn 0.24.1. All experiments are run on GPUs (either NVIDIA Tesla V100-SXM3-32GB-H or NVIDIA A100-SXM4-40GB), while usually distributing training over multiple GPUs. Memory in the order of 20 GB is required for training depending on batch size; as an example with 4 GPUs and a batch size of 16 (stacks of four frames), around 18 GB were utilized.

Metrics are gathered after 200 gradient steps each, to allow for comparisons over training, and training is stopped after 5000 steps. A batch size of 16 is used consistently. If the batch size is modified we recommend scaling all gradient-step-depended hyperparameters like the number of permitted gradient steps, the annealing for the $z^{pres}$ and $z^{scale}$ (see SPACE \cite{SPACE2020}) and the learning rate using approximately the classic square root scaling \cite{krizhevsky2014one}. For better comparability, we increase the learning rate also for SPACE and show, in Appendix~\ref{appendix:origin}, that this modification does not impede the performance of our baseline.

For evaluation of variance and continuity, we train each model on 5 seeds and provide the mean and standard deviation for each metric. This principle is repeated for 8 Atari 2600 games.

In practice, we used local mode computation solely with the game Tennis, where it was necessary due to changing backgrounds (when switching sides), and otherwise computed a global mode image. For efficiency, we stop collecting images for the global mode image early and simply modify the image if the final global mode image was obvious (at least regarding moving objects).

Furthermore, no extensive hyperparameter search was conducted, but values were mostly inspired by the originally chosen values of SPACE and spot-checked for modifications.

\section{Exact Numerical Results}
\label{appendix:numerical_results}
In this section, we provide tables of the empirical evaluation we have conducted to allow for future comparisons.
\subsection{Object Detection}
\label{appendix:results_ap}
{

This section expands on the evaluation of Object Detection by providing alternative views of the performance. First, we expand the set of metrics to accuracy and AP, and second also display the performance when considering all objects, including the otherwise filtered irrelevant objects. These metrics are shown in the Tables \ref{appendix:table_object_detection_first} to \ref{appendix:table_object_detection_last}. 

The results mostly match the insights gained before, with naturally slightly worse performance with all objects and low absolute AP explained by how SPACE tends to use larger bounding boxes. 

}

\setlength{\tabcolsep}{8pt}
\begin{table}[ht]
\centering
\begin{tabular}{|c||c|c|c|c|c|c|c|c|}
\hline
\multicolumn{4}{|c|}{\textbf{Accuracy (Relevant)}} \\
\hline
\hline
\textbf{Configuration} & \textbf{SPACE} & \textbf{SP.+MOC(w/o OC)} & \textbf{SP.+MOC } \\
\hline
\textbf{Air Raid} & 0.1 $\pm$ 0.1 & \textbf{63.4 $\pm$ 4.9} & \textbf{67.4 $\pm$ 2.9} \\
\textbf{Boxing} & 21.4 $\pm$ 14.4 & \textbf{77.2 $\pm$ 1.0} & \textbf{77.1 $\pm$ 1.5} \\
\textbf{Carnival} & 1.9 $\pm$ 2.0 & \textbf{40.9 $\pm$ 3.0} & \textbf{41.3 $\pm$ 1.1} \\
\textbf{Ms. Pacman} & 0.0 $\pm$ 0.0 & \textbf{55.7 $\pm$ 1.0} & \textbf{55.5 $\pm$ 1.3} \\
\textbf{Pong} & 0.0 $\pm$ 0.0 & \textbf{89.9 $\pm$ 0.4} & 66.7 $\pm$ 6.3 \\
\textbf{Riverraid} & 3.5 $\pm$ 3.1 & \textbf{30.6 $\pm$ 1.4} & \textbf{32.1 $\pm$ 1.7} \\
\textbf{Space Invaders} & \textbf{11.4 $\pm$ 14.6} & \textbf{11.3 $\pm$ 5.2} & \textbf{11.1 $\pm$ 1.2} \\
\textbf{Tennis} & 0.0 $\pm$ 0.0 & \textbf{32.1 $\pm$ 14.7} & \textbf{32.6 $\pm$ 15.1} \\
\hline
\end{tabular}
\caption{Accuracy (share of frames, where the number of objects is correct) of relevant objects.}
\label{appendix:table_object_detection_first}
\end{table}

\begin{table}[ht]
\centering
\begin{tabular}{|c||c|c|c|c|c|c|c|c|}
\hline
\multicolumn{4}{|c|}{\textbf{Average AP (Relevant)}} \\
\hline
\hline
\textbf{Configuration} & \textbf{SPACE} & \textbf{SP.+MOC(w/o OC)} & \textbf{SP.+MOC } \\
\hline
\textbf{Air Raid} & 22.6 $\pm$ 4.7 & \textbf{67.9 $\pm$ 0.6} & \textbf{67.4 $\pm$ 2.2} \\
\textbf{Boxing} & 12.5 $\pm$ 1.8 & 59.7 $\pm$ 2.3 & \textbf{62.5 $\pm$ 2.4} \\
\textbf{Carnival} & 20.5 $\pm$ 5.4 & \textbf{42.9 $\pm$ 1.3} & 42.1 $\pm$ 2.8 \\
\textbf{Ms. Pacman} & 0.1 $\pm$ 0.1 & \textbf{30.6 $\pm$ 0.8} & 29.1 $\pm$ 0.9 \\
\textbf{Pong} & 0.3 $\pm$ 0.4 & \textbf{16.8 $\pm$ 1.0} & \textbf{17.4 $\pm$ 0.6} \\
\textbf{Riverraid} & 16.9 $\pm$ 1.9 & \textbf{28.5 $\pm$ 2.0} & 28.2 $\pm$ 0.4 \\
\textbf{Space Invaders} & 29.2 $\pm$ 4.8 & \textbf{34.5 $\pm$ 2.7} & \textbf{35.5 $\pm$ 2.0} \\
\textbf{Tennis} & 0.0 $\pm$ 0.0 & \textbf{14.0 $\pm$ 2.0} & \textbf{14.7 $\pm$ 2.6} \\
\hline
\end{tabular}
\caption{Average AP on relevant objects.}
\label{table:relevant_ap_avg}
\end{table}

\begin{table}[ht]
\centering
\begin{tabular}{|c||c|c|c|c|c|c|c|c|}
\hline
\multicolumn{4}{|c|}{\textbf{F-Score (Relevant)}} \\
\hline
\hline
\textbf{Configuration} & \textbf{SPACE} & \textbf{SP.+MOC(w/o OC)} & \textbf{SP.+MOC } \\
\hline
\textbf{Air Raid} & 51.3 $\pm$ 3.2 & 89.7 $\pm$ 1.2 & \textbf{91.5 $\pm$ 1.0} \\
\textbf{Boxing} & 58.8 $\pm$ 4.6 & \textbf{91.8 $\pm$ 0.4} & \textbf{92.1 $\pm$ 0.7} \\
\textbf{Carnival} & 51.6 $\pm$ 7.8 & \textbf{93.1 $\pm$ 1.5} & \textbf{93.0 $\pm$ 1.2} \\
\textbf{Ms. Pacman} & 0.5 $\pm$ 1.0 & \textbf{91.4 $\pm$ 0.3} & \textbf{91.5 $\pm$ 0.4} \\
\textbf{Pong} & 0.2 $\pm$ 0.3 & \textbf{97.6 $\pm$ 0.2} & 92.8 $\pm$ 1.3 \\
\textbf{Riverraid} & 65.0 $\pm$ 1.2 & \textbf{68.6 $\pm$ 1.1} & \textbf{69.1 $\pm$ 1.0} \\
\textbf{Space Invaders} & \textbf{86.5 $\pm$ 7.4} & \textbf{86.3 $\pm$ 2.6} & \textbf{87.2 $\pm$ 0.3} \\
\textbf{Tennis} & 0.0 $\pm$ 0.0 & \textbf{62.7 $\pm$ 6.3} & \textbf{62.6 $\pm$ 8.0} \\
\hline
\end{tabular}
\caption{$F_1$ on relevant objects. Repeated from the evaluation for comparison.}
\label{table:relevant_f_score}
\end{table}

\begin{table}[ht]
\centering
\begin{tabular}{|c||c|c|c|c|c|c|c|c|}
\hline
\multicolumn{4}{|c|}{\textbf{Accuracy (All)}} \\
\hline
\hline
\textbf{Configuration} & \textbf{SPACE} & \textbf{SP.+MOC(w/o OC)} & \textbf{SP.+MOC } \\
\hline
\textbf{Air Raid} & 1.6 $\pm$ 2.1 & \textbf{36.5 $\pm$ 3.2} & 27.3 $\pm$ 3.9 \\
\textbf{Boxing} & 8.3 $\pm$ 18.6 & \textbf{22.7 $\pm$ 1.6} & \textbf{23.2 $\pm$ 1.0} \\
\textbf{Carnival} & 1.3 $\pm$ 1.5 & \textbf{32.3 $\pm$ 2.7} & \textbf{30.1 $\pm$ 3.3} \\
\textbf{Ms. Pacman} & 0.0 $\pm$ 0.1 & \textbf{31.0 $\pm$ 4.8} & \textbf{33.9 $\pm$ 1.7} \\
\textbf{Pong} & 0.0 $\pm$ 0.0 & \textbf{46.3 $\pm$ 0.2} & 34.5 $\pm$ 2.4 \\
\textbf{Riverraid} & 0.0 $\pm$ 0.0 & 3.2 $\pm$ 1.9 & \textbf{3.4 $\pm$ 1.8} \\
\textbf{Space Invaders} & \textbf{8.4 $\pm$ 7.3} & \textbf{5.8 $\pm$ 3.3} & \textbf{4.4 $\pm$ 0.6} \\
\textbf{Tennis} & 0.0 $\pm$ 0.0 & 0.0 $\pm$ 0.0 & 0.0 $\pm$ 0.0 \\
\hline
\end{tabular}
\caption{Accuracy on all objects.}
\label{table:all_accuracy}
\end{table}

\begin{table}[ht]
\centering
\begin{tabular}{|c||c|c|c|c|c|c|c|c|}
\hline
\multicolumn{4}{|c|}{\textbf{Average AP (All)}} \\
\hline
\hline
\textbf{Configuration} & \textbf{SPACE} & \textbf{SP.+MOC(w/o OC)} & \textbf{SP.+MOC } \\
\hline
\textbf{Air Raid} & 19.2 $\pm$ 3.0 & \textbf{63.1 $\pm$ 0.9} & \textbf{62.7 $\pm$ 1.7} \\
\textbf{Boxing} & 5.0 $\pm$ 1.0 & \textbf{19.1 $\pm$ 2.6} & \textbf{19.5 $\pm$ 2.9} \\
\textbf{Carnival} & 18.5 $\pm$ 4.4 & \textbf{42.5 $\pm$ 1.3} & 41.6 $\pm$ 2.3 \\
\textbf{Ms. Pacman} & 0.1 $\pm$ 0.1 & \textbf{30.1 $\pm$ 1.9} & \textbf{30.6 $\pm$ 1.4} \\
\textbf{Pong} & 8.3 $\pm$ 10.0 & 23.6 $\pm$ 1.0 & \textbf{26.9 $\pm$ 2.8} \\
\textbf{Riverraid} & 11.4 $\pm$ 1.0 & 21.8 $\pm$ 0.8 & \textbf{22.4 $\pm$ 0.3} \\
\textbf{Space Invaders} & 28.1 $\pm$ 3.7 & \textbf{32.7 $\pm$ 2.3} & \textbf{33.2 $\pm$ 2.0} \\
\textbf{Tennis} & 0.0 $\pm$ 0.0 & \textbf{8.3 $\pm$ 1.2} & \textbf{8.6 $\pm$ 1.5} \\
\hline
\end{tabular}
\caption{Average AP on all objects.}
\label{table:all_ap_avg}
\end{table}

\begin{table}[ht]
\centering
\begin{tabular}{|c||c|c|c|c|c|c|c|c|}
\hline
\multicolumn{4}{|c|}{\textbf{F-Score (All)}} \\
\hline
\hline
\textbf{Configuration} & \textbf{SPACE} & \textbf{SP.+MOC(w/o OC)} & \textbf{SP.+MOC } \\
\hline
\textbf{Air Raid} & 49.1 $\pm$ 4.3 & \textbf{87.7 $\pm$ 0.3} & \textbf{87.9 $\pm$ 0.9} \\
\textbf{Boxing} & 24.5 $\pm$ 2.8 & \textbf{70.4 $\pm$ 0.8} & \textbf{70.5 $\pm$ 0.8} \\
\textbf{Carnival} & 48.6 $\pm$ 9.7 & \textbf{90.6 $\pm$ 1.2} & \textbf{90.6 $\pm$ 0.7} \\
\textbf{Ms. Pacman} & 0.4 $\pm$ 0.8 & \textbf{88.6 $\pm$ 2.5} & \textbf{90.5 $\pm$ 0.4} \\
\textbf{Pong} & 10.7 $\pm$ 13.2 & \textbf{91.5 $\pm$ 0.1} & 87.4 $\pm$ 0.9 \\
\textbf{Riverraid} & 45.0 $\pm$ 0.7 & \textbf{77.1 $\pm$ 0.8} & \textbf{76.6 $\pm$ 0.7} \\
\textbf{Space Invaders} & \textbf{87.5 $\pm$ 4.1} & \textbf{85.1 $\pm$ 1.1} & \textbf{85.2 $\pm$ 0.3} \\
\textbf{Tennis} & 3.6 $\pm$ 8.1 & \textbf{40.5 $\pm$ 5.3} & \textbf{40.2 $\pm$ 6.1} \\
\hline
\end{tabular}
\caption{$F_1$ on all objects.}
\label{appendix:table_object_detection_last}
\end{table}

\subsection{Object Continuity}
\label{appendix:object_consistency}
{

Similar to the results in object detection we can report the result on all (instead of only the relevant) objects. Furthermore, we provide the accuracy of all few-shot learners with their standard deviation: Namely trained with 1, 4, 16, and 64 encoding per class and the cluster neighbor-based classifier. These metrics are shown in the Tables \ref{appendix:table_object_consistency_first} to \ref{appendix:table_object_consistency_last}. 

The data similarly shows the advantage of using object continuity and reveals the interesting relation between more training data and higher accuracy, i.e. the knowledge intrinsic to a larger sample. The effect of object continuity is decreasing with more training data, as with little data the better spread in the encoding space is very important, while with more data even correlations present with baselines can be approximated more precisely.

The concludes the presentation of further results from the evaluation.

}
\FloatBarrier

\setlength{\tabcolsep}{8pt}
\begin{table}[htbp]
\centering
\begin{tabular}{|c||c|c|c|c|c|c|c|c|}
\hline
\multicolumn{4}{|c|}{\textbf{Relevant Few-Shot Accuracy (trained with 1)}} \\
\hline
\hline
\textbf{Configuration} & \textbf{SPACE} & \textbf{SP.+MOC(w/o OC)} & \textbf{SP.+MOC } \\
\hline
\textbf{Air Raid} & 45.8 $\pm$ 11.0 & 58.0 $\pm$ 4.5 & \textbf{59.5 $\pm$ 0.5} \\
\textbf{Boxing} & 51.8 $\pm$ 26.1 & 80.4 $\pm$ 6.2 & \textbf{95.4 $\pm$ 0.8} \\
\textbf{Carnival} & 0.0 $\pm$ 0.0 & 46.1 $\pm$ 4.0 & \textbf{48.1 $\pm$ 4.8} \\
\textbf{Ms. Pacman} & 0.0 $\pm$ 0.0 & 32.9 $\pm$ 4.2 & \textbf{65.9 $\pm$ 6.4} \\
\textbf{Pong} & 0.0 $\pm$ 0.0 & 68.0 $\pm$ 6.2 & \textbf{83.7 $\pm$ 13.8} \\
\textbf{Riverraid} & 51.5 $\pm$ 6.6 & 34.0 $\pm$ 7.7 & \textbf{61.4 $\pm$ 1.2} \\
\textbf{Space Invaders} & 0.0 $\pm$ 0.0 & 29.4 $\pm$ 4.7 & \textbf{63.2 $\pm$ 4.0} \\
\textbf{Tennis} & 0.0 $\pm$ 0.0 & 34.6 $\pm$ 7.0 & \textbf{38.7 $\pm$ 7.7} \\
\hline
\end{tabular}
\caption{Relevant Accuracy of a linear model trained with 1 sample per class.}
\label{appendix:table_object_consistency_first}
\end{table}

\begin{table}[htbp]
\centering
\begin{tabular}{|c||c|c|c|c|c|c|c|c|}
\hline
\multicolumn{4}{|c|}{\textbf{Relevant Few-Shot Accuracy (trained with 4)}} \\
\hline
\hline
\textbf{Configuration} & \textbf{SPACE} & \textbf{SP.+MOC(w/o OC)} & \textbf{SP.+MOC } \\
\hline
\textbf{Air Raid} & 57.7 $\pm$ 21.1 & 58.3 $\pm$ 5.8 & \textbf{66.9 $\pm$ 5.3} \\
\textbf{Boxing} & 46.8 $\pm$ 25.1 & 93.2 $\pm$ 2.9 & \textbf{96.7 $\pm$ 0.6} \\
\textbf{Carnival} & 0.0 $\pm$ 0.0 & 48.7 $\pm$ 5.5 & \textbf{70.2 $\pm$ 5.0} \\
\textbf{Ms. Pacman} & 2.5 $\pm$ 5.6 & 35.7 $\pm$ 4.4 & \textbf{77.6 $\pm$ 1.8} \\
\textbf{Pong} & 0.0 $\pm$ 0.0 & 75.0 $\pm$ 7.2 & \textbf{98.3 $\pm$ 0.5} \\
\textbf{Riverraid} & 60.1 $\pm$ 6.1 & 47.3 $\pm$ 7.9 & \textbf{78.3 $\pm$ 1.1} \\
\textbf{Space Invaders} & 0.0 $\pm$ 0.0 & 40.3 $\pm$ 4.9 & \textbf{60.2 $\pm$ 1.8} \\
\textbf{Tennis} & 0.0 $\pm$ 0.0 & 35.3 $\pm$ 4.1 & \textbf{40.1 $\pm$ 4.8} \\
\hline
\end{tabular}
\caption{Relevant Accuracy of a linear model trained with 4 samples per class.}
\label{appendix:table:relevant_few_shot_accuracy_with_4}
\end{table}

\begin{table}[htbp]
\centering
\begin{tabular}{|c||c|c|c|c|c|c|c|c|}
\hline
\multicolumn{4}{|c|}{\textbf{Relevant Few-Shot Accuracy (trained with 16)}} \\
\hline
\hline
\textbf{Configuration} & \textbf{SPACE} & \textbf{SP.+MOC(w/o OC)} & \textbf{SP.+MOC } \\
\hline
\textbf{Air Raid} & 69.7 $\pm$ 22.0 & 75.3 $\pm$ 5.3 & \textbf{83.1 $\pm$ 2.6} \\
\textbf{Boxing} & 60.0 $\pm$ 17.2 & 90.2 $\pm$ 3.1 & \textbf{94.9 $\pm$ 0.9} \\
\textbf{Carnival} & 0.0 $\pm$ 0.0 & \textbf{76.7 $\pm$ 1.9} & \textbf{77.0 $\pm$ 2.2} \\
\textbf{Ms. Pacman} & 3.3 $\pm$ 7.5 & 62.5 $\pm$ 2.5 & \textbf{84.1 $\pm$ 1.3} \\
\textbf{Pong} & 0.0 $\pm$ 0.0 & 85.4 $\pm$ 5.6 & \textbf{98.6 $\pm$ 0.2} \\
\textbf{Riverraid} & 59.9 $\pm$ 6.8 & 61.4 $\pm$ 4.0 & \textbf{80.9 $\pm$ 3.1} \\
\textbf{Space Invaders} & 0.0 $\pm$ 0.0 & \textbf{70.0 $\pm$ 3.8} & \textbf{66.5 $\pm$ 1.9} \\
\textbf{Tennis} & 0.0 $\pm$ 0.0 & 34.0 $\pm$ 7.3 & \textbf{41.0 $\pm$ 2.2} \\
\hline
\end{tabular}
\caption{Relevant Accuracy of a linear model trained with 16 samples per class.}
\label{table:relevant_few_shot_accuracy_with_16}
\end{table}

\begin{table}[htbp]
\centering
\begin{tabular}{|c||c|c|c|c|c|c|c|c|}
\hline
\multicolumn{4}{|c|}{\textbf{Relevant Few-Shot Accuracy (trained with 64)}} \\
\hline
\hline
\textbf{Configuration} & \textbf{SPACE} & \textbf{SP.+MOC(w/o OC)} & \textbf{SP.+MOC } \\
\hline
\textbf{Air Raid} & \textbf{78.0 $\pm$ 26.8} & \textbf{85.6 $\pm$ 1.4} & \textbf{84.4 $\pm$ 1.3} \\
\textbf{Boxing} & 57.6 $\pm$ 19.0 & \textbf{94.7 $\pm$ 1.4} & \textbf{95.2 $\pm$ 1.3} \\
\textbf{Carnival} & 0.0 $\pm$ 0.0 & \textbf{96.5 $\pm$ 0.8} & \textbf{97.4 $\pm$ 1.6} \\
\textbf{Ms. Pacman} & 0.8 $\pm$ 1.9 & 79.5 $\pm$ 2.1 & \textbf{92.0 $\pm$ 1.3} \\
\textbf{Pong} & 0.0 $\pm$ 0.0 & 95.7 $\pm$ 1.7 & \textbf{98.7 $\pm$ 0.4} \\
\textbf{Riverraid} & 69.7 $\pm$ 6.6 & 75.8 $\pm$ 2.3 & \textbf{87.8 $\pm$ 0.7} \\
\textbf{Space Invaders} & 0.0 $\pm$ 0.0 & \textbf{75.7 $\pm$ 2.3} & 70.6 $\pm$ 1.9 \\
\textbf{Tennis} & 0.0 $\pm$ 0.0 & 42.3 $\pm$ 12.8 & \textbf{59.3 $\pm$ 4.7} \\
\hline
\end{tabular}
\caption{Relevant Accuracy of a linear model trained with 64 samples per class.}
\label{table:relevant_few_shot_accuracy_with_64}
\end{table}

\begin{table}[htbp]
\centering
\begin{tabular}{|c||c|c|c|c|c|c|c|c|}
\hline
\multicolumn{4}{|c|}{\textbf{Relevant Cluster Classifier Accuracy}} \\
\hline
\hline
\textbf{Configuration} & \textbf{SPACE} & \textbf{SP.+MOC(w/o OC)} & \textbf{SP.+MOC } \\
\hline
\textbf{Air Raid} & 70.5 $\pm$ 23.8 & \textbf{83.6 $\pm$ 1.5} & \textbf{81.4 $\pm$ 4.9} \\
\textbf{Boxing} & 57.3 $\pm$ 31.2 & 92.5 $\pm$ 1.8 & \textbf{96.4 $\pm$ 0.5} \\
\textbf{Carnival} & 0.0 $\pm$ 0.0 & \textbf{93.5 $\pm$ 1.1} & 86.7 $\pm$ 3.3 \\
\textbf{Ms. Pacman} & 0.8 $\pm$ 1.9 & 62.6 $\pm$ 6.6 & \textbf{84.1 $\pm$ 4.0} \\
\textbf{Pong} & 0.0 $\pm$ 0.0 & 94.0 $\pm$ 2.4 & \textbf{98.7 $\pm$ 0.4} \\
\textbf{Riverraid} & \textbf{79.1 $\pm$ 13.2} & 65.8 $\pm$ 5.9 & 66.2 $\pm$ 0.9 \\
\textbf{Space Invaders} & 0.0 $\pm$ 0.0 & \textbf{78.2 $\pm$ 2.2} & \textbf{80.1 $\pm$ 3.6} \\
\textbf{Tennis} & 0.0 $\pm$ 0.0 & 42.1 $\pm$ 8.1 & \textbf{50.6 $\pm$ 6.2} \\
\hline
\end{tabular}
\caption{Relevant Accuracy of the model classifying by label of closest cluster mean.}
\label{table:relevant_few_shot_accuracy_cluster_nn}
\end{table}

\begin{table}[htbp]
\centering
\begin{tabular}{|c||c|c|c|c|c|c|c|c|}
\hline
\multicolumn{4}{|c|}{\textbf{Few-Shot Accuracy (all classes, trained with 1)}} \\
\hline
\hline
\textbf{Configuration} & \textbf{SPACE} & \textbf{SP.+MOC(w/o OC)} & \textbf{SP.+MOC } \\
\hline
\textbf{Air Raid} & 30.0 $\pm$ 15.1 & 49.7 $\pm$ 5.8 & \textbf{57.3 $\pm$ 4.0} \\
\textbf{Boxing} & 45.3 $\pm$ 15.0 & 43.1 $\pm$ 10.5 & \textbf{73.9 $\pm$ 5.0} \\
\textbf{Carnival} & 0.0 $\pm$ 0.0 & 38.6 $\pm$ 3.5 & \textbf{47.4 $\pm$ 3.8} \\
\textbf{Ms. Pacman} & 0.0 $\pm$ 0.0 & 20.9 $\pm$ 2.9 & \textbf{63.8 $\pm$ 4.6} \\
\textbf{Pong} & 0.0 $\pm$ 0.0 & 47.4 $\pm$ 2.7 & \textbf{66.3 $\pm$ 4.4} \\
\textbf{Riverraid} & 28.3 $\pm$ 9.6 & 22.5 $\pm$ 4.2 & \textbf{49.0 $\pm$ 1.9} \\
\textbf{Space Invaders} & 0.0 $\pm$ 0.0 & 24.5 $\pm$ 5.4 & \textbf{54.9 $\pm$ 3.0} \\
\textbf{Tennis} & 0.0 $\pm$ 0.0 & 25.3 $\pm$ 7.7 & \textbf{30.1 $\pm$ 4.7} \\
\hline
\end{tabular}
\caption{Accuracy on all objects of a linear model trained with 1 sample per class.}
\label{table:all_few_shot_accuracy_with_1}
\end{table}

\begin{table}[htbp]
\centering
\begin{tabular}{|c||c|c|c|c|c|c|c|c|}
\hline
\multicolumn{4}{|c|}{\textbf{Few-Shot Accuracy (all classes, trained with 4)}} \\
\hline
\hline
\textbf{Configuration} & \textbf{SPACE} & \textbf{SP.+MOC(w/o OC)} & \textbf{SP.+MOC } \\
\hline
\textbf{Air Raid} & 39.0 $\pm$ 9.7 & 50.8 $\pm$ 3.1 & \textbf{61.6 $\pm$ 5.1} \\
\textbf{Boxing} & 51.5 $\pm$ 12.7 & 50.7 $\pm$ 8.3 & \textbf{75.5 $\pm$ 2.6} \\
\textbf{Carnival} & 0.0 $\pm$ 0.0 & 43.5 $\pm$ 5.4 & \textbf{63.4 $\pm$ 5.6} \\
\textbf{Ms. Pacman} & 2.1 $\pm$ 4.8 & 30.5 $\pm$ 5.5 & \textbf{70.0 $\pm$ 2.1} \\
\textbf{Pong} & 0.0 $\pm$ 0.0 & 49.4 $\pm$ 4.7 & \textbf{69.6 $\pm$ 8.0} \\
\textbf{Riverraid} & 27.8 $\pm$ 7.6 & 24.6 $\pm$ 4.3 & \textbf{61.0 $\pm$ 1.0} \\
\textbf{Space Invaders} & 0.0 $\pm$ 0.0 & 34.4 $\pm$ 4.1 & \textbf{52.4 $\pm$ 2.7} \\
\textbf{Tennis} & 0.0 $\pm$ 0.0 & 26.6 $\pm$ 7.3 & \textbf{31.1 $\pm$ 4.6} \\
\hline
\end{tabular}
\caption{Accuracy on all objects of a linear model trained with 4 samples per class.}
\label{table:all_few_shot_accuracy_with_4}
\end{table}

\begin{table}[htbp]
\centering
\begin{tabular}{|c||c|c|c|c|c|c|c|c|}
\hline
\multicolumn{4}{|c|}{\textbf{Few-Shot Accuracy (all classes, trained with 16)}} \\
\hline
\hline
\textbf{Configuration} & \textbf{SPACE} & \textbf{SP.+MOC(w/o OC)} & \textbf{SP.+MOC } \\
\hline
\textbf{Air Raid} & 40.0 $\pm$ 11.0 & 67.3 $\pm$ 8.1 & \textbf{74.5 $\pm$ 5.6} \\
\textbf{Boxing} & 53.0 $\pm$ 10.9 & 54.9 $\pm$ 9.7 & \textbf{75.3 $\pm$ 2.2} \\
\textbf{Carnival} & 0.0 $\pm$ 0.0 & \textbf{73.0 $\pm$ 4.1} & \textbf{74.2 $\pm$ 3.6} \\
\textbf{Ms. Pacman} & 2.1 $\pm$ 4.8 & 53.2 $\pm$ 4.4 & \textbf{74.5 $\pm$ 2.5} \\
\textbf{Pong} & 0.0 $\pm$ 0.0 & 67.2 $\pm$ 2.2 & \textbf{77.9 $\pm$ 7.3} \\
\textbf{Riverraid} & 45.8 $\pm$ 6.6 & 45.8 $\pm$ 3.9 & \textbf{71.7 $\pm$ 2.4} \\
\textbf{Space Invaders} & 0.0 $\pm$ 0.0 & \textbf{62.4 $\pm$ 5.4} & 56.3 $\pm$ 2.9 \\
\textbf{Tennis} & 0.0 $\pm$ 0.0 & 26.2 $\pm$ 4.8 & \textbf{32.1 $\pm$ 2.6} \\
\hline
\end{tabular}
\caption{Accuracy on all objects of a linear model trained with 16 samples per class.}
\label{table:all_few_shot_accuracy_with_16}
\end{table}

\begin{table}[htbp]
\centering
\begin{tabular}{|c||c|c|c|c|c|c|c|c|}
\hline
\multicolumn{4}{|c|}{\textbf{Few-Shot Accuracy (all classes, trained with 64)}} \\
\hline
\hline
\textbf{Configuration} & \textbf{SPACE} & \textbf{SP.+MOC(w/o OC)} & \textbf{SP.+MOC } \\
\hline
\textbf{Air Raid} & 45.2 $\pm$ 18.3 & 78.1 $\pm$ 4.9 & \textbf{80.1 $\pm$ 4.7} \\
\textbf{Boxing} & 67.4 $\pm$ 14.4 & 56.4 $\pm$ 10.4 & \textbf{77.3 $\pm$ 2.6} \\
\textbf{Carnival} & 0.0 $\pm$ 0.0 & \textbf{89.7 $\pm$ 2.1} & \textbf{90.5 $\pm$ 1.4} \\
\textbf{Ms. Pacman} & 6.4 $\pm$ 14.3 & 66.3 $\pm$ 4.9 & \textbf{78.6 $\pm$ 2.7} \\
\textbf{Pong} & 0.0 $\pm$ 0.0 & 75.8 $\pm$ 1.6 & \textbf{84.4 $\pm$ 4.7} \\
\textbf{Riverraid} & 60.0 $\pm$ 12.5 & 53.8 $\pm$ 2.9 & \textbf{79.4 $\pm$ 1.4} \\
\textbf{Space Invaders} & 0.0 $\pm$ 0.0 & \textbf{68.1 $\pm$ 4.3} & 59.0 $\pm$ 2.2 \\
\textbf{Tennis} & 0.0 $\pm$ 0.0 & 33.8 $\pm$ 8.8 & \textbf{43.0 $\pm$ 3.8} \\
\hline
\end{tabular}
\caption{Accuracy on all objects of a linear model trained with 64 samples per class.}
\label{table:all_few_shot_accuracy_with_64}
\end{table}

\begin{table}[htbp]
\centering
\begin{tabular}{|c||c|c|c|c|c|c|c|c|}
\hline
\multicolumn{4}{|c|}{\textbf{Cluster Classifier Accuracy (all)}} \\
\hline
\hline
\textbf{Configuration} & \textbf{SPACE} & \textbf{SP.+MOC(w/o OC)} & \textbf{SP.+MOC } \\
\hline
\textbf{Air Raid} & 63.5 $\pm$ 25.1 & \textbf{77.7 $\pm$ 2.5} & \textbf{75.0 $\pm$ 4.4} \\
\textbf{Boxing} & 79.0 $\pm$ 9.6 & 58.0 $\pm$ 9.1 & \textbf{85.6 $\pm$ 3.5} \\
\textbf{Carnival} & 0.0 $\pm$ 0.0 & \textbf{84.9 $\pm$ 6.2} & 76.4 $\pm$ 1.8 \\
\textbf{Ms. Pacman} & 19.8 $\pm$ 44.2 & 63.5 $\pm$ 6.6 & \textbf{84.3 $\pm$ 6.9} \\
\textbf{Pong} & 0.0 $\pm$ 0.0 & \textbf{72.1 $\pm$ 7.7} & 71.2 $\pm$ 10.9 \\
\textbf{Riverraid} & \textbf{81.3 $\pm$ 6.1} & 50.0 $\pm$ 4.1 & 74.0 $\pm$ 4.5 \\
\textbf{Space Invaders} & 0.0 $\pm$ 0.0 & \textbf{72.2 $\pm$ 3.5} & \textbf{71.0 $\pm$ 8.7} \\
\textbf{Tennis} & 0.0 $\pm$ 0.0 & 38.8 $\pm$ 12.0 & \textbf{40.9 $\pm$ 12.3} \\
\hline
\end{tabular}
\caption{Accuracy on all objects of the model classifying by label of closest cluster mean.}
\label{appendix:table_object_consistency_last}
\end{table}
\FloatBarrier

\clearpage
\newpage
\subsection{Playing Atari game results}
\label{app:playing_games_results}
We here give detailed numerical results for each seed of the different playing agents. These correspond to final scores of agents initially trained with perfect information extracted from the RAM, then evaluated with object detection based on the perfect information baseline (RAM) as well as SPACE models, SPACE + MOC w/o OC, and SPACE with our complete SPACE + MOC architecture. The best deep agent is bolded. We use for each reported \textit{seed}, the object detector model trained with this seed, as well as the few-shot ($16$) classification model trained with the same seed, and initialize the environment based on this seed.

\begin{table}[ht]
\centering
\small
{\def\arraystretch{.9}
\setlength{\tabcolsep}{5pt}
\begin{tabular}{@{}r|c|c|ccc@{}}
\toprule
\scriptsize Seed & \multicolumn{1}{c|}{Random}       & \multicolumn{1}{c|}{Info* }         & \multicolumn{1}{c}{SPACE}         & \scriptsize SP.+MOC w/o OC         & \multicolumn{1}{r}{SP.+MOC}                              \\ \midrule
0    & $\text{-}2.9\mbox{\scriptsize$\pm10.6$}$ & $28.4\mbox{\scriptsize$\pm14.6$}$ & $\text{-}3.7\mbox{\scriptsize$\pm5.8$}$ & $\textbf{18.1}\mbox{\scriptsize$\pm17.1$}$ & $14.6\mbox{\scriptsize$\pm9.5$}$  \\
1    & $\text{-}0.8\mbox{\scriptsize$\pm4.9$}$  & $\textbf{52.1}\mbox{\scriptsize$\pm3.5$}$  & $\text{-}3.7\mbox{\scriptsize$\pm8.0$}$ & $\text{-}0.3\mbox{\scriptsize$\pm7.3$}$  & $\text{-}2.1\mbox{\scriptsize$\pm6.1$}$  \\
2    & $\text{-}1.3\mbox{\scriptsize$\pm4.7$}$  & $43.9\mbox{\scriptsize$\pm2.2$}$  & $\text{-}5.6\mbox{\scriptsize$\pm5.9$}$ & $5.4\mbox{\scriptsize$\pm5.9$}$   & $9.3\mbox{\scriptsize$\pm8.1$}$   \\
3    & $\textbf{2.6}\mbox{\scriptsize$\pm3.3$}$   & $35.8\mbox{\scriptsize$\pm13.7$}$ & $\text{-}3.5\mbox{\scriptsize$\pm7.4$}$ & $0.6\mbox{\scriptsize$\pm6.8$}$   & $\textbf{21.9}\mbox{\scriptsize$\pm14.5$}$ \\
4    & $\text{-}0.1\mbox{\scriptsize$\pm3.6$}$  & $8.8\mbox{\scriptsize$\pm19.9$}$  & $\textbf{3.8}\mbox{\scriptsize$\pm6.5$}$  & $\text{-}0.9\mbox{\scriptsize$\pm9.3$}$  & $\text{-}1.9\mbox{\scriptsize$\pm8.0$}$  \\ \midrule
avg & $\text{-}0.5\mbox{\scriptsize$\pm2. $ } $ & $36\mbox{\scriptsize$\pm 17$ }$ & \text{-}$3.5\mbox{\scriptsize$\pm 4.3$}$ & $4.7\mbox{\scriptsize$\pm7.8$ } $ & $\textbf{8.4}\mbox{\scriptsize $\pm9.9$ } $ \\ \bottomrule
\end{tabular}
}
%
%
%
\centering
\small
{\def\arraystretch{.9}
\setlength{\tabcolsep}{5pt}
\begin{tabular}{@{}c|c|ccc@{}}
\toprule
\multicolumn{1}{c|}{Random}       & \multicolumn{1}{c|}{Info* }         & \multicolumn{1}{c}{SPACE}         & \scriptsize SP.+MOC w/o OC         & \multicolumn{1}{r}{SP.+MOC}       \\ \midrule
$\text{-}20.8\mbox{\scriptsize$\pm0.4$}$ & $\textbf{21.0}\mbox{\scriptsize$\pm0.0$}$ & $\text{-}21.0^*$  & $\text{-}20.0\mbox{\scriptsize$\pm1.0$}$ & $\text{-}20.9\mbox{\scriptsize$\pm0.3$}$ \\
$\text{-}20.6\mbox{\scriptsize$\pm0.6$}$ & $\textbf{21.0}\mbox{\scriptsize$\pm0.0$}$ & $\text{-}\textbf{19.9}\mbox{\scriptsize$\pm0.8$}$ & $\text{-}14.0\mbox{\scriptsize$\pm4.1$}$ & $\text{-}18.3\mbox{\scriptsize$\pm2.1$}$ \\
$\text{-}\textbf{20.1}\mbox{\scriptsize$\pm0.7$}$ & $18.0\mbox{\scriptsize$\pm0.0$}$ & $\text{-}21.0\mbox{\scriptsize$\pm0.0$}$ & $\text{-}16.3\mbox{\scriptsize$\pm3.2$}$ & $\textbf{4.8}\mbox{\scriptsize$\pm11.3$}$  \\
$\text{-}20.7\mbox{\scriptsize$\pm0.6$}$ & $\textbf{21.0}\mbox{\scriptsize$\pm0.0$}$ & $\text{-}21.0^*$  & $\text{-}19.3\mbox{\scriptsize$\pm1.8$}$ & $\text{-}19.5\mbox{\scriptsize$\pm0.8$}$ \\
$\text{-}20.5\mbox{\scriptsize$\pm0.5$}$ & $20.0\mbox{\scriptsize$\pm0.0$}$ & $\text{-}20.5\mbox{\scriptsize$\pm0.8$}$ & $\text{-}\textbf{9.2}\mbox{\scriptsize$\pm10.2$}$ & $\text{-}2.5\mbox{\scriptsize$\pm14.9$}$ \\ \midrule
\text{-}$20.5\mbox{\scriptsize$\pm0.3$} $ & $20\mbox{\scriptsize$\pm1.3$}$ & \text{-}$21\mbox{\scriptsize$\pm 0.5$ }$ & \text{-}$16\mbox{\scriptsize$\pm 4.4$} $ & \text{-}$\textbf{11}\mbox{\scriptsize$\pm 12$}$ \\ \bottomrule
\end{tabular}
}
\caption{Different object-centric agents playing boxing (\textbf{Left}), and Pong (\textbf{Right}). Average obtained scores for each seeded agent, with the corresponding seed, as well as their average scores. For each seed, we ran random agents, agents with perfect information obtained from the RAM, as well as the original SPACE baseline model, SPACE + MOC and the ablated SPACE + MOC without object continuity. For the SPACE baseline, agent with a ``*'' correspond to models that detect no object.}
\end{table}

\section{Atari-OC -- Per game details}
\label{appendix:ocatari_detail}
Following, we list our method to detect objects for creating the Atari-OC dataset, for each individual Atari game we used. We provide a ``relevant'' tag that stands for the relevance of the object to successfully play the game. Scores and lives, usually part of the HUD for humans, are classified as non-relevant. We also give an estimate of the quality of the labeling method, by listing precision and recall that were obtained via a set of 16 randomly chosen and hand-labeled images of each game. 

\begin{table}[ht]\centering
\small
\begin{tabular}{@{}lccrr@{}}
\toprule
\textbf{Object/Entity} & \textbf{Method} & \textbf{Relevant} & \textbf{Precision} & \textbf{Recall} \\ 
\midrule
player & Position + Size & Yes & 1.000 & 1.000 \\ 
score & Position & No & 1.000 & 1.000 \\ 
building & Position + Size & Yes & 1.000 & 1.000 \\ 
shot & Position + Size & Yes & 1.000 & 1.000 \\ 
enemy & Color + Exclusion Principle & Yes & 1.000 & 1.000 \\
\bottomrule
\end{tabular}
\caption{Air Raid}
\end{table}

\begin{table}[ht]\centering
\small
\begin{tabular}{@{}lccrr@{}}
\toprule
\textbf{Object/Entity} & \textbf{Method} & \textbf{Relevant} & \textbf{Precision} & \textbf{Recall} \\ 
\midrule
black & Color + Position & Yes & 1.000 & 1.000 \\ 
black\_score & Fixed & No & 1.000 & 0.656 \\ 
clock & Fixed & No & 1.000 & 1.000 \\ 
white & Color + Position & Yes & 1.000 & 1.000 \\ 
white\_score & Fixed & No & 1.000 & 0.656 \\ 
logo & Fixed & No & 1.000 & 1.000 \\
\bottomrule
\end{tabular}
\caption{Boxing}
\end{table}

\begin{table}[ht]\centering
\small
\begin{tabular}{@{}lccrr@{}}
\toprule
\textbf{Object/Entity} & \textbf{Method} & \textbf{Relevant} & \textbf{Precision} & \textbf{Recall} \\ 
\midrule
pacman & Color & Yes & 1.000 & 1.000 \\ 
sue & Color & Yes & 1.000 & 1.000 \\ 
inky & Color & Yes & 1.000 & 1.000 \\ 
pinky & Color & Yes & 1.000 & 1.000 \\ 
blinky & Color & Yes & 1.000 & 1.000 \\ 
eyes & Size + Relevant Label & Yes & 1.000 & 1.000 \\ 
blue\_ghost & Color & Yes & 1.000 & 1.000 \\ 
white\_ghost & Color & Yes & 1.000 & 1.000 \\ 
fruit & Color & Yes & 1.000 & 1.000 \\ 
level & Color & No & 1.000 & 1.000 \\ 
life1 & Color & No & 1.000 & 1.000 \\ 
life2 & Color & No & 1.000 & 1.000 \\ 
score & Color & No & 1.000 & 1.000 \\ 
power\_pill & Color & Yes & 1.000 & 1.000 \\
\bottomrule
\end{tabular}
\caption{Ms. Pacman}
\end{table}

\begin{table}[ht]\centering
\small
\begin{tabular}{@{}lccrr@{}}
\toprule
\textbf{Object/Entity} & \textbf{Method} & \textbf{Relevant} & \textbf{Precision} & \textbf{Recall} \\ 
\midrule
owl & Color & Yes & 0.769 & 1.000 \\ 
rabbit & Color + Size & Yes & 1.000 & 0.971 \\ 
shooter & Color & Yes & 1.000 & 1.000 \\ 
refill & Color + Size & Yes & 0.917 & 1.000 \\ 
bonus & Color & No & n/a & 0.000 \\ 
duck & Color + Size & Yes &  1.000 & 1.000 \\ 
flying\_duck & Color + Size & Yes & 1.000 & 1.000 \\ 
score & Color + Position & No & 1.000 & 1.000 \\ 
pipes & Fixed & No & 1.000 & 1.000 \\ 
eating\_duck & Color + Position & Yes & 1.000 & 1.000 \\ 
bullet & Color & Yes & 1.000 & 1.000 \\
\bottomrule
\end{tabular}
\caption{Carnival}
\end{table}

\begin{table}[ht]\centering
\small
\begin{tabular}{@{}lccrr@{}}
\toprule
\textbf{Object/Entity} & \textbf{Method} & \textbf{Relevant} & \textbf{Precision} & \textbf{Recall} \\ 
\midrule
player & Color + Position & Yes & 1.000 & 1.000 \\ 
enemy & Color + Position & Yes & 1.000 & 1.000 \\ 
ball & Color + Extra Weight & Yes & 1.000 & 0.933 \\ 
enemy\_score & Fixed & No & 1.000 & 1.000 \\ 
player\_score & Fixed & No & 1.000 & 0.750 \\
\bottomrule
\end{tabular}
\caption{Pong}
\end{table}

\begin{table}[ht]\centering
\small
\begin{tabular}{@{}lccrr@{}}
\toprule
\textbf{Object/Entity} & \textbf{Method} & \textbf{Relevant} & \textbf{Precision} & \textbf{Recall} \\ 
\midrule
player & Color + Position + Size & Yes & 0.778 & 1.000 \\ 
fuel\_gauge & Color + Position + Size & Yes & 1.000 & 1.000 \\ 
fuel & Color + Position + Size & Yes & 0.432 & 1.000 \\ 
lives & Color + Position + Size & No & 1.000 & 1.000 \\ 
logo & Color + Position + Size & No & 1.000 & 1.000 \\ 
score & Color + Position + Size & No & 1.000 & 1.000 \\ 
shot & Color + Position + Size & No & 1.000 & 1.000 \\ 
fuel\_board & Fixed & No & 1.000 & 1.000 \\ 
building & Color + Position + Size & No & n/a & 0.000 \\ 
street & Color + Position + Size & Yes & 1.000 & 1.000 \\ 
enemy & Color + Position + Size & Yes & 1.000 & 1.000 \\
\bottomrule
\end{tabular}
\caption{Riverraid}
\end{table}

\begin{table}[ht]\centering
\small
\begin{tabular}{@{}lccrr@{}}
\toprule
\textbf{Object/Entity} & \textbf{Method} & \textbf{Relevant} & \textbf{Precision} & \textbf{Recall} \\ 
\midrule
left\_score & Fixed & No & 1.000 & 1.000 \\ 
right\_score & Fixed & No & 1.000 & 1.000 \\ 
enemy\_0 & Color + Position & Yes & 1.000 & 1.000 \\ 
enemy\_1 & Color + Position & Yes & 1.000 & 1.000 \\ 
enemy\_2 & Color + Position & Yes & 1.000 & 1.000 \\ 
enemy\_3 & Color + Position & Yes & 1.000 & 1.000 \\ 
enemy\_4 & Color + Position & Yes & 1.000 & 1.000 \\ 
enemy\_5 & Color + Position & Yes & 1.000 & 1.000 \\ 
space\_ship & Color & Yes & 1.000 & 1.000 \\ 
player & Color & Yes & 1.000 & 1.000 \\ 
block & Color & Yes & 1.000 & 1.000 \\ 
bullet & Color & Yes & 1.000 & 1.000 \\
\bottomrule
\end{tabular}
\caption{Space Invaders}
\end{table}

\begin{table}[ht]\centering
\small
\begin{tabular}{@{}lccrr@{}}
\toprule
\textbf{Object/Entity} & \textbf{Method} & \textbf{Relevant} & \textbf{Precision} & \textbf{Recall} \\ 
\midrule
player & Color & Yes & 1.000 & 1.000 \\ 
enemy & Color & Yes & 1.000 & 1.000 \\ 
ball & Color & Yes & 1.000 & 1.000 \\ 
ball\_shadow & Color & Yes & 1.000 & 1.000 \\ 
logo & Color & No & 1.000 & 1.000 \\ 
player\_score & Color & No & 1.000 & 0.500 \\ 
enemy\_score & Color & No & 1.000 & 0.750 \\
\bottomrule
\end{tabular}
\caption{Tennis}
\caption{Tables describing the heuristic method for semi-automatic object labeling. We also report the precision and recall of this heuristic labeling approach.}
\label{appendix::table::labeling}

\end{table}


\end{document}